\documentclass{article}

\usepackage{iclr2027_conference,times}

\usepackage{amsmath}
\usepackage{etoolbox}
\BeforeBeginEnvironment{equation}{\begingroup\small}
\AfterEndEnvironment{equation}{\endgroup}
\BeforeBeginEnvironment{equation*}{\begingroup\small}
\AfterEndEnvironment{equation*}{\endgroup}
\usepackage{amssymb}
\usepackage{booktabs}
\usepackage{graphicx}
\usepackage{subcaption}

\usepackage[table]{xcolor}
\usepackage{array}
\usepackage{tabularx}
\usepackage{multirow}
\usepackage{microtype}
\usepackage{listings}
\usepackage{float}
\usepackage{wrapfig}
\usepackage{needspace}
\BeforeBeginEnvironment{wraptable}{\Needspace{15\baselineskip}}
\usepackage{capt-of}
\usepackage{hyperref}
\hypersetup{breaklinks=true}
\usepackage{CJKutf8}
\usepackage{tikz}
\usepackage{pgfplots}
\usetikzlibrary{arrows.meta}
\pgfplotsset{compat=1.18}

\graphicspath{{figures/}}
\definecolor{oursblue}{HTML}{2F6B9A}
\definecolor{oursbluepale}{HTML}{EAF2F8}
\definecolor{baselinegray}{HTML}{7F8C8D}
\definecolor{breakred}{HTML}{C44E52}
\definecolor{shiftorange}{HTML}{DD8452}
\definecolor{equivblue}{HTML}{4C72B0}
\newcommand{\best}[1]{\textbf{#1}}

\newcolumntype{Y}{>{\centering\arraybackslash}X}
\makeatletter
\patchcmd{\@maketitle}{\vskip 0.3in minus 0.1in}{\vskip 0.12in minus 0.03in}{}{\PackageError{arxiv-layout}{Title spacing patch failed}{Check the ICLR style.}}
\makeatother
\newcommand{\compactmainspacing}{%
  \setlength{\parskip}{3.5pt plus 0.5pt minus 0.5pt}%
  \setlength{\textfloatsep}{10pt plus 2pt minus 2pt}%
  \setlength{\floatsep}{8pt plus 2pt minus 2pt}%
  \setlength{\intextsep}{8pt plus 2pt minus 2pt}%
  \captionsetup{skip=4pt}%
}
\title{From Input to Output: A Flexible Agent for Dual-End Interpretation of Sparse Autoencoder Features}

\iclrfinalcopy
\author{%
  \parbox[t]{\dimexpr\textwidth-2\tabcolsep\relax}{%
    \centering\normalfont
    \textbf{Dewen Liu}\textsuperscript{1,2}\thanks{Equal contribution; order determined by random coin flip.}\textsuperscript{\phantom{*},}\setcounter{footnote}{3}\thanks{Work done during internship at Tsinghua University.}
    \quad \textbf{Zixuan Li}\textsuperscript{1}\footnotemark[1]
    \quad \textbf{Jonathan Pan}\textsuperscript{1,3}
    \quad \textbf{Zhao Wu}\textsuperscript{1}\\[0.15em]
    \textbf{Zijun Yao}\textsuperscript{1}
    \quad \textbf{Juanzi Li}\textsuperscript{1}
    \quad \textbf{Xiaozhi Wang}\textsuperscript{1}\\[0.4em]
    \textsuperscript{1}Tsinghua University
    \qquad \textsuperscript{2}Fudan University
    \qquad \textsuperscript{3}University of Edinburgh\\[0.4em]
    {\small\texttt{dwliu23@m.fudan.edu.cn}\quad
    \texttt{zx-li21@mails.tsinghua.edu.cn}\\[0.1em]
    \texttt{xzwang@sz.tsinghua.edu.cn}}%
  }%
}

\begin{document}
\begin{CJK*}{UTF8}{gbsn}
\raggedbottom

\maketitle
\fancyhead{}

\begin{abstract}

Sparse autoencoders (SAEs) are an important tool for mechanistic interpretability, but interpreting their many features remains challenging. Existing methods characterize input-side activation patterns and output-side intervention effects, yet often leave their functional connection implicit, while input-side evidence collection typically relies on costly large-corpus scans. We introduce \emph{functional interpretation}, which characterizes an SAE feature as a mapping from its activating input semantics to its output effects under intervention, and present Dual-End Agentic Feature Interpretation (DAFI), an agent that actively gathers evidence and refines input-side, output-side, and functional interpretations through component-specific feedback. Its short-context token probing enables on-demand activation evidence collection without a full corpus scan. On GemmaScope, DAFI improves Input score by 13.1 percentage points over SAGE and Output score by 38.9 points over Token Change, while being substantially more token-efficient than a general-purpose coding agent. Skills distilled from successful refinements raise the held-out joint pass rate from 58.0\% to 92.0\% and improve both interpretation quality and efficiency when transferred to a new model--SAE setting. Across features with reliable endpoint interpretations, 70.7\% exhibit non-equivalent input and output semantics. On AxBench, DAFI also improves steering-feature selection over output-score filtering. Code is available at \url{https://github.com/THUAIS-Lab/DAFI}.
\end{abstract}

\begingroup
\compactmainspacing
\section{Introduction}

\begin{wrapfigure}{r}{0.56\textwidth}
    \vspace{-0.75\baselineskip}
    \centering
    \includegraphics[width=\linewidth]{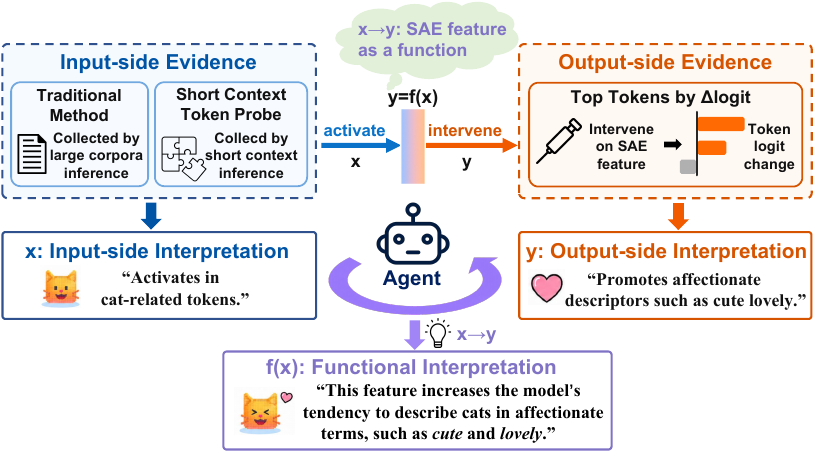}
    \caption{Overview of DAFI's three-component interpretation target.}
    \label{fig:framework}
    \vspace{-0.50\baselineskip}
\end{wrapfigure}

Sparse autoencoders (SAEs) map dense model activations into a higher-dimensional sparse feature space, yielding tens of thousands of fine-grained features per dictionary for mechanistic analysis \citep{elhage2022superposition,huben2024sparse,bricken2023monosemanticity,gao2025scaling,lieberum2024gemmascope,deng2026qwenscope}. How to reliably interpret these features remains an open problem. Existing automated methods interpret those features from two endpoints: input-side methods characterize activation contexts \citep{bills2023language,paulo2025automatically,maher2026multishot,han2026sage}; output-side methods characterize causal effects on model outputs \citep{paulo2025automatically,gur2025outputcentric}. Neither endpoint fully characterizes a feature. For example, a feature may activate on cat-related tokens but promote affectionate descriptors such as \emph{cute} and \emph{lovely} when intervened on. Labeling it as either a ``cat-related feature'' or a ``cute-related feature'' captures only one endpoint. Viewed from input to output, the feature maps cat-related activation contexts to affectionate output effects. This explanation describes the feature's function during model inference. We refer to such an input--output explanation as a functional interpretation (Figure~\ref{fig:framework}). Existing input-side methods also scan a fixed large corpus for activation evidence. These scans amortize across many features but are inefficient for small feature sets.

Reliable feature interpretation involves heterogeneous evidence and requires different refinement operations across the three components \citep{ma2025revising,maher2026multishot,han2026sage}. The next operation depends on the evidence and diagnostic feedback, so the refinement trajectory cannot be fixed in advance. An agent selects the next operation accordingly.

To address these problems, we make the following contributions:

{\setlength{\leftmargini}{1.5em}
\begin{itemize}
    \item We formulate SAE feature interpretation as a three-component target comprising input-side, output-side, and functional interpretations, with component-specific diagnostics and an overall criterion requiring all three components to pass. Using an Equivalent--Shift--Break taxonomy, we find that most features with reliable endpoint interpretations fall into the Shift or Break categories, underscoring the need for functional interpretation to \mbox{determine whether and how the two endpoints are connected.}

    \item We introduce Dual-End Agentic Feature Interpretation (DAFI), an agent that actively acquires evidence and selectively refines failed components, and validate it across two LLM--SAE settings. On matched GemmaScope features, DAFI outperforms component-matched input- and output-side baselines while additionally producing functional interpretations.

    \item We equip DAFI with tools and a self-evolving mechanism to support on-demand evidence collection and improvement across features. For targeted features, its short-context token probing collects input-side evidence on demand without requiring a full corpus scan, while achieving input-side scores comparable to those obtained from corpus-derived activation examples. DAFI self-evolves by distilling refinements into reusable skills that improve interpretation quality and efficiency on unseen features and transfer across LLM--SAE settings.

\end{itemize}
}

\section{Related Work}

\paragraph{Sparse autoencoders}

Sparse autoencoders (SAEs) learn a sparse, higher-dimensional representation of dense LLM activations: an encoder maps each activation to sparse feature coefficients, and a decoder reconstructs the activation from those features, providing finer-grained units than polysemantic neurons \citep{elhage2022superposition,huben2024sparse,bricken2023monosemanticity,gao2025scaling,rajamanoharan2024jumping}. Interpreting these features helps us understand model mechanisms \citep{bricken2023monosemanticity,huben2024sparse,jing2025lingualens,marks2025sparsecircuits}. We aim to provide more complete interpretations of SAE features and more effective methods for producing them.

\paragraph{Automated interpretation}

Existing automated methods follow two main approaches: input-side and output-side interpretation. Input-side methods interpret features based on activating examples, with recent agentic variants iteratively refining their interpretations \citep{bills2023language,huben2024sparse,paulo2025automatically,maher2026multishot,han2026sage}. Collecting this evidence relies on a full corpus scan, which is costly when interpreting a few features of interest \citep{gur2025outputcentric}. Output-side methods use vocabulary projections or feature interventions to characterize changes in model outputs, but lack fine-grained diagnostics for iteratively refining output-side hypotheses \citep{paulo2025automatically,gur2025outputcentric}. Prior work either interprets a feature from a single endpoint or simply combines evidence from both endpoints into a single feature interpretation \citep{gur2025outputcentric}. We instead introduce a functional interpretation that characterizes a feature as a mapping from its input-side activation pattern to its causal output effects. DAFI evaluates all three interpretations separately and uses component-specific feedback to guide refinement, while short-context token \mbox{probing enables on-demand input-side evidence collection.}

\paragraph{Self-evolving agents}
Building on early reasoning-and-action and self-reflection agents \citep{yao2023react,shinn2023reflexion}, recent self-evolving systems improve across tasks by accumulating and reusing experience \citep{wang2024voyager,zhao2024expel,hu2025automated,yin2025godel,zheng2025skillweaver,zhang2026darwin}. \Needspace{4\baselineskip}Such experience can be retained at different levels of abstraction, from concrete cases to transferable strategies \citep{suzgun2026dynamic,zhang2026ace}. DAFI retains successful revisions as case skills and distills recurring patterns \mbox{into general skills for interpreting subsequent features.}

\section{Method}

DAFI evaluates input-side, output-side, and functional interpretations, uses their diagnostics to guide an agent loop, and distills successful revisions into reusable skills (Figure~\ref{fig:agent-overview}).

\subsection{Problem Formulation: Endpoint and Relational Interpretations}
\label{sec:dual-end-functional-interpretation}

Activation-based methods explain what inputs activate a feature, with agentic variants refining these descriptions through activation feedback \citep{bills2023language,paulo2025automatically,han2026sage}. Output-centric methods characterize downstream effects, and combining input and output descriptions improves faithfulness \citep{gur2025outputcentric}. We extend this view by treating the functional connection between the endpoints as a distinct, testable interpretation target: how a feature links its activating conditions to the output behavior it promotes.

Figure~\ref{fig:agent-overview} illustrates a feature activated by \emph{cat} and \emph{kitten} that promotes \emph{cute} and \emph{lovely} when amplified. The functional interpretation is that the feature preferentially responds to cat-related semantics, and amplifying this activation biases the model toward affectionate expression.

Given a language model, an SAE, and a target feature $i$, we seek a hypothesis triplet
\[
H_t(i)=\left(h_t^{\mathrm{in}},h_t^{\mathrm{out}},h_t^c\right),
\]
where $t$ is the refinement round. The input hypothesis $h_t^{\mathrm{in}}$ describes the semantic concepts that preferentially activate the feature. The output hypothesis $h_t^{\mathrm{out}}$ describes the semantic bias induced in the model's output by intervening on the feature. The functional hypothesis $h_t^c$ explains the feature's role in linking its activating semantics to this intervention-induced output bias. These components are evaluated separately to guide targeted refinement.

\begin{figure*}[!t]
    \centering
    \includegraphics[width=\textwidth]{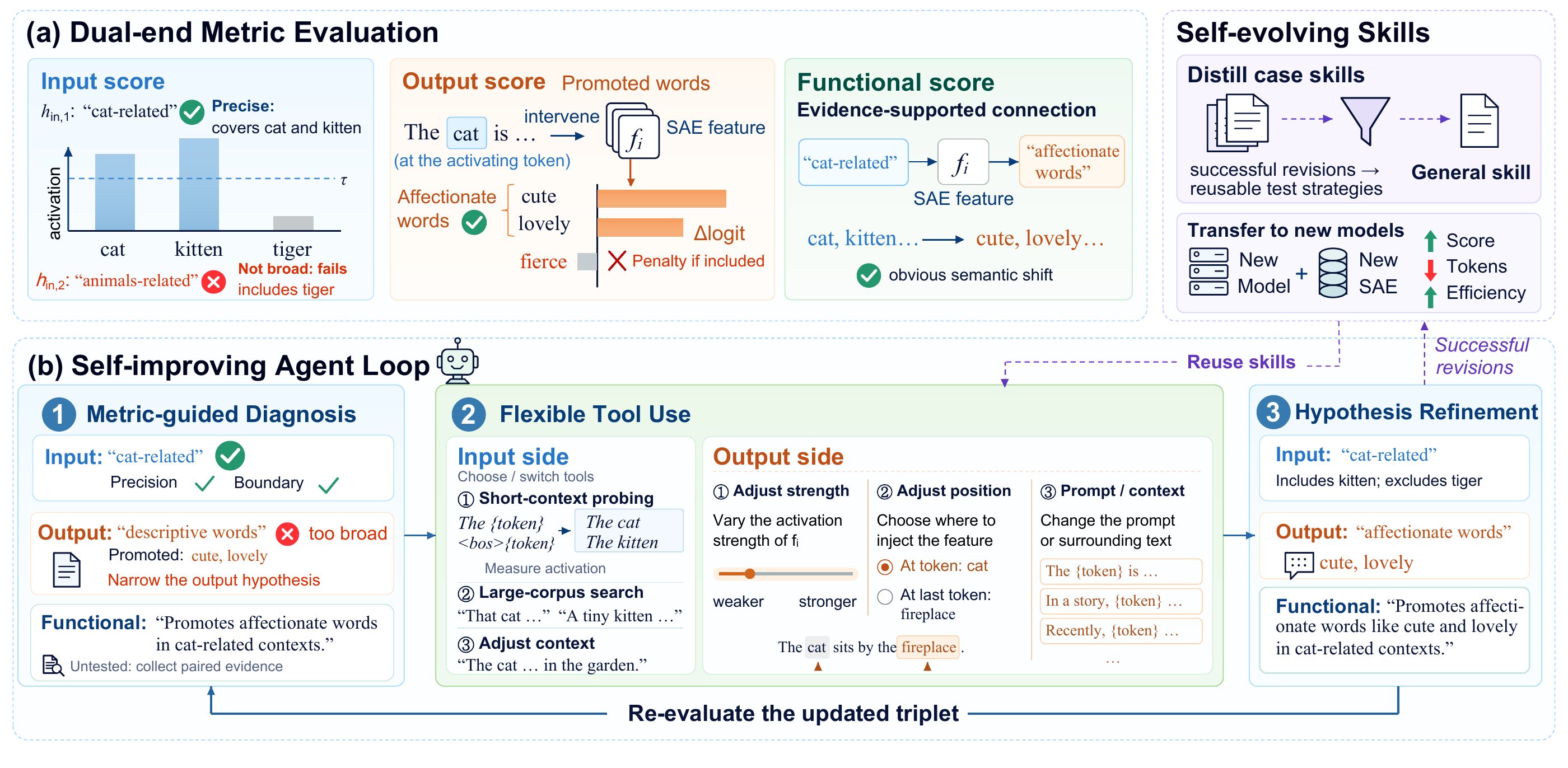}
    \caption{DAFI: (a) three interpretation metrics; (b) the agent loop with reusable skills.}
    \label{fig:agent-overview}
\end{figure*}

\subsection{Metric Design}
\label{sec:component-evaluation}

We evaluate the hypothesis triplet using three scores. The \textbf{Input score} measures whether the input hypothesis captures the feature's activation preferences and semantic boundaries. The \textbf{Output score} measures whether the output hypothesis captures the semantic bias induced by feature intervention. The \textbf{Functional score} assesses whether the activating input semantics and intervention-induced output semantics have a coherent semantic relation, and whether the functional hypothesis accurately describes this relation using evidence from both endpoints.

\paragraph{Input score.}
An informative input interpretation must identify reliable triggers and delimit their semantic scope. We evaluate activation coverage and boundary rejection, following \citet{han2026sage} and \citet{ma2025revising}, respectively. Given $h_t^{\mathrm{in}}$, an LLM constructs positive prompts $P$ that should activate the feature and semantically adjacent boundary prompts $B$ that should not. With feature activation $f_i(x)$ on prompt $x$, we compute
\begin{equation}
\begin{aligned}
\mathrm{Act}(i)&=\frac{1}{|P|}\sum_{x\in P}\mathbf{1}[f_i(x)>\tau],\\
\mathrm{Bnd}(i)&=\frac{1}{|B|}\sum_{x\in B}\mathbf{1}[f_i(x)\leq\tau],
\end{aligned}
\end{equation}
where $\tau$ is the dynamically computed activation threshold for the current hypothesis. Act tests whether inputs covered by the hypothesis actually activate the feature. Bnd tests whether nearby inputs outside its stated scope remain inactive. We define Input score as their harmonic mean,
\begin{equation}
S_{\mathrm{in}}(i)=\frac{2\,\mathrm{Act}(i)\,\mathrm{Bnd}(i)}{\mathrm{Act}(i)+\mathrm{Bnd}(i)},
\end{equation}
with $S_{\mathrm{in}}(i)=0$ when both components are zero. For aggregate results, we apply the same formula to the mean Act and mean Bnd over the reported feature set. The harmonic mean penalizes an interpretation that performs well on only one component. Act, Bnd, and $S_{\mathrm{in}}$ range from 0 to 1. For the stricter input-side pass criterion, both Act and Bnd must individually reach 0.8.

\paragraph{Output score.}
An output interpretation should capture the token patterns promoted by feature intervention while excluding those it suppresses. Let $T_i^+$ and $T_i^-$ contain the retained tokens with positive and negative logit changes. An LLM judge identifies tokens matching $h_t^{\mathrm{out}}$, giving
\begin{equation}
S_{\mathrm{out}}(i)=\mathrm{Coverage}(h_t^{\mathrm{out}},T_i^+)
-\mathrm{Penalty}(h_t^{\mathrm{out}},T_i^-).
\end{equation}
Coverage rewards positive logit-change mass captured by the hypothesis. Penalty subtracts negative-change magnitudes for tokens it incorrectly describes as promoted, discouraging broad explanations. In Figure~\ref{fig:agent-overview}a, \emph{affectionate words} covers \emph{cute} and \emph{lovely}, whereas including suppressed \emph{fierce} incurs a penalty. Both terms are normalized by total positive mass. The score passes at 0.5 and becomes negative when penalty exceeds coverage; Appendix~\ref{sec:m2-definition} provides details.

\paragraph{Functional score.}
An LLM judge tests whether $h_t^c$ explains the feature's functional role by connecting the semantics of its activating inputs and promoted outputs. In the running example, the target is the supported semantic shift from cat-related contexts to affectionate outputs. The judge grounds this relation in the endpoint hypotheses, activating contexts $C_i$, and intervention evidence $E_i$:
\begin{equation}
S_{\mathrm{func}}(i)=J_{\mathrm{LLM}}\left(
h_t^{\mathrm{in}},C_i,h_t^{\mathrm{out}},E_i,h_t^c
\right)\in\{1,\ldots,5\}.
\end{equation}
Scores range from 1 (no supported relation) to 5 (strongly supported relation). Passing requires at least 4, indicating a clear semantic connection supported by both endpoints (Appendix~\ref{sec:explanation-and-judge-templates}).

\subsection{Agent Design}
\label{sec:agent-design}

Figure~\ref{fig:agent-overview}b presents DAFI's self-improving agent loop, which combines metric-guided diagnosis, flexible tool use, and hypothesis refinement. The agent repeatedly evaluates and updates the hypothesis triplet, while successful revisions are distilled into reusable skills that guide subsequent interpretations.

\paragraph{Self-improving agent loop.}
DAFI investigates a feature's role during generation by examining its output effects in the contexts that activate it. Input-side evidence guides the choice of intervention contexts and token positions, while the resulting output changes help assess whether and how the activating semantics relate to the promoted semantics. The functional hypothesis thus provides an explicit relation to investigate through coordinated probing and intervention.

Starting from an initial hypothesis triplet, the agent reasons over metric feedback and the underlying evidence to decide which hypotheses to revise and what to test next. It can refine a description using existing observations or adjust probing and intervention conditions to obtain further evidence. When the input--output relation remains unclear, these choices allow the agent to investigate the connection and revise the functional hypothesis based on newly observed effects. The agent re-evaluates the updated triplet and repeats this process until all criteria pass or refinement stops, returning the best-supported complete triplet.

\paragraph{Flexible tool use.}
The agent carries out refinements using configurable evidence-collection tools.

On the input side, it can retrieve corpus-derived activating examples from Neuronpedia\footnote{\url{https://www.neuronpedia.org/}} or use \emph{short-context token probing}. This tool inserts candidate vocabulary tokens into configurable templates, such as \verb|The {token}| and \verb|<bos>{token}|, allowing full-vocabulary probing with short inputs for a target feature. The agent can adjust templates and surrounding context to identify activating conditions and refine the input hypothesis.

On the output side, the agent can intervene in activating contexts, targeting the maximally activating token by default. It can adjust intervention strength, token position, and prompt context, then inspect positive and negative output-token logit changes. These observations identify the output semantics promoted or suppressed under the selected conditions, guiding subsequent interventions and revisions to the output and functional hypotheses.

\paragraph{Self-evolving skills.}
Successful revisions are stored as case skills; recurring diagnostic and repair patterns are distilled into general skills. These skills guide later hypotheses, tool choices, probing contexts, intervention settings, and rerun scope, allowing effective evidence-acquisition and refinement strategies to be reused across features. The reusable-skills branch in Figure~\ref{fig:agent-overview} feeds this experience into subsequent runs and supports transfer across language model and SAE settings.

\section{Experiments}

Our experiments evaluate whether DAFI can interpret what activates an SAE feature, how it affects model output, and the functional connection between these endpoints at a practical computational cost. We compare interpretation quality and cost with existing methods, examine individual design choices, and test whether accumulated experience transfers to unseen features and a new LLM--SAE setting. For features with well-supported endpoint interpretations, we classify \mbox{their semantic relations as Equivalent, Shift, or Break.}

\subsection{Experiment Setting}
DAFI generates initial input-side, output-side, and functional hypotheses for each target feature and refines them for up to 10 rounds. We evaluate the three components using the Input, Output, and Functional scores, respectively, as defined in Section~\ref{sec:component-evaluation}. Input score is the harmonic mean of activation coverage and boundary rejection. The input-side component passes only if both underlying components are at least 0.8; the output-side and functional components pass if Output score is at least 0.5 and Functional score is at least 4, respectively. Joint success \mbox{requires the selected complete triplet to satisfy all criteria.}

For the main experiments, we use GemmaScope \citep{lieberum2024gemmascope}, a suite of pretrained sparse autoencoders for Gemma 2, specifically the \texttt{gemmascope-res-16k} residual-stream SAEs for Gemma-2-2B \citep{gemmateam2024gemma2}. We sample equal numbers of features from layers 0, 6, 12, 18, and 24. For the cross-setting transfer experiment, we use Qwen-Scope \citep{deng2026qwenscope} W32K residual-stream SAEs using Top-$k$ activation with $k=50$ for Qwen3-1.7B-Base \citep{yang2025qwen3}, sampling 20 features from each of layers 0, 7, 14, 21, and 27.

We use DeepSeek-V4-Pro \citep{deepseek2026v4} as the backbone model for hypothesis generation, refinement, and primary automated evaluation, with temperature set to 0. To avoid potential same-model evaluation bias \citep{panickssery2024llm,xu2024pride,chen2025beyond}, we independently re-evaluate all eligible output-side and functional hypotheses from the final traces of the 250-feature GemmaScope experiment using GPT-5.6-Sol \citep{openai2026gpt56sol}. GPT-5.6-Sol receives the same evidence and scoring rules as DeepSeek-V4-Pro, but not DeepSeek-V4-Pro's original scores or rationales. To assess agreement with human judgments, two researchers with experience in SAE interpretability independently rate functional interpretation quality on a stratified 40-feature sample while blinded to the system scores and to each other's labels. \mbox{Appendices~\ref{sec:cross-model-judge-consistency} and~\ref{sec:human-validation} report the respective protocols and results.}

\subsection{Main Results and Evaluation Reliability}
\label{sec:efficiency}
\label{sec:framework-evaluation}
We compare DAFI's interpretation quality and efficiency with component-matched and end-to-end baselines. The evaluation scores the three components of each interpretation triplet separately: Input score summarizes activation coverage and boundary rejection using their harmonic mean, Output score evaluates the output-side interpretation based on observed intervention-induced token changes, and Functional score evaluates the functional relation between the two endpoints. The appendix reports the two Input score components separately.

\paragraph{Baseline comparisons.}

Because existing SAE interpretation methods do not directly produce the full triplet of input-side, output-side, and functional interpretations, we use both component-matched and end-to-end baselines. Each DAFI run produces the complete triplet. For component-matched comparisons, we compare its input-side component with SAGE \citep{han2026sage} and its output-side component with Token Change \mbox{\citep{gur2025outputcentric}} on 100 features. For the end-to-end comparison, we use Claude Code \citep{anthropic2026claudecode} as a baseline, prompting it to generate hypotheses for all three components following basic diagnostic instructions specified in \texttt{CLAUDE.md}. We evaluate both a budget-restricted configuration, whose round-start budget is set by DAFI's per-feature usage, and unrestricted execution. Because of Claude Code's high execution cost, both configurations use 25 features and are marked with an asterisk in Table~\ref{tab:eafi-baseline-comparison}.

\begin{table}[!htbp]
\centering
\footnotesize
\setlength{\tabcolsep}{4pt}
\begin{tabular}{@{}lcccc@{}}
\toprule
\multirow{2}{*}{Method} & Input & Output & Functional & Tokens \\
& score (\%) & score (\%) & score & (M) $\downarrow$ \\
\midrule
SAGE & 78.3 & \textemdash & \textemdash & 0.796 \\
Token Change & \textemdash & 27.8 & \textemdash & 0.00484 \\
\multicolumn{5}{@{}l}{Claude Code} \\
\quad Restricted$^{*}$ & 84.0 & 35.0 & 2.8 & 3.082 \\
\quad Unrestricted$^{*}$ & 94.0 & 54.8 & 4.0 & 13.296 \\
\midrule
\rowcolor{oursbluepale}
DAFI & \best{91.4} & \best{66.7} & \best{4.0} & \best{0.758} \\
\bottomrule
\end{tabular}
\caption{Baseline comparison. Input score is the harmonic mean of activation coverage and boundary rejection. Input and Output scores are percentage-scaled, whereas Functional score uses a 1--5 scale. Costs are millions of tokens per feature. $^{*}$ denotes the 25-feature Claude Code evaluation; all other rows use 100 features. Dashes: inapplicable metrics. Blue: DAFI. $\downarrow$: lower is better.}
\label{tab:eafi-baseline-comparison}
\end{table}

Against the two component-matched baselines on 100 features, DAFI achieves a 13.1-point higher Input score than SAGE (91.4\% versus 78.3\%). DAFI also uses fewer tokens per feature (0.758M versus 0.796M), while additionally producing output-side and functional interpretations. Compared with Token Change, DAFI's iterative refinement achieves a higher Output score (66.7\% versus 27.8\%), with higher and more consistent scores across layers. Appendix~\ref{sec:budget-details} reports the Input score components and detailed results for the Claude Code configurations. Table~\ref{tab:m2-layerwise-token-change} in Appendix~\ref{sec:output-centric-validation} reports the layer-wise output comparison.   Layer-wise refinement results and analysis are reported in Appendix~\ref{sec:layer-level-notes}.

\paragraph{Cross-model and human consistency.}
Because the Output and Functional scores contain LLM-judged components, we first assess their cross-model consistency; for the Functional score, we additionally compare the scores with human judgments. GPT-5.6-Sol agrees with DeepSeek-V4-Pro on 87.1\% of Output score and 80.4\% of Functional score pass decisions; Appendix~\ref{sec:cross-model-judge-consistency} reports the full results. In a separate human evaluation, two researchers with experience in SAE-based interpretability independently review a sample of 40 features. Functional score correlates with their ratings of functional interpretation quality (Spearman $\rho=0.809$ and $0.777$), with binary agreement of 95.0\% and 92.5\%, respectively. These results support agreement between Functional score and human judgments of functional interpretation quality; Appendix~\ref{sec:human-validation} reports the full statistics.

\subsection{Design Analysis}

We next evaluate three design elements: short-context probing as an alternative evidence source, skill accumulation through sequential interpretation, and skill transfer across LLM--SAE settings.

\paragraph{Short-context token probing as an alternative evidence source.}
\begin{wraptable}{r}{0.48\textwidth}
\vspace{-0.75\baselineskip}
\centering
\footnotesize
\setlength{\tabcolsep}{2pt}
\begin{tabular*}{\linewidth}{@{\extracolsep{\fill}}lccc@{}}
\toprule
Evidence & Initial & Refined & Gain \\
\midrule
Neuronpedia & \best{82.1} & \best{91.4} & +9.3 \\
\rowcolor{oursbluepale}
Short-context probing & 72.7 & 88.5 & \best{+15.8} \\
\bottomrule
\end{tabular*}
\caption{Input scores before and after adaptive refinement with Neuronpedia or one-template short-context probing on 100 matched GemmaScope features. Initial and refined scores are reported as percentages; gain is their absolute difference in percentage points. Bold values are best; blue marks short-context probing.}
\label{tab:probing-vs-neuronpedia}
\vspace{-0.50\baselineskip}
\end{wraptable}
Corpus scanning can amortize inference across many features, but it is costly for small feature sets and limited by the coverage of a fixed dataset. We assess short-context token probing as an alternative source of initial evidence on 100 GemmaScope features, starting from a single \verb|<bos>{token}| template. After refinement, probing achieves a larger Input score gain than Neuronpedia initialization (15.8 versus 9.3 percentage points), although its final Input score remains lower (88.5\% versus 91.4\%; Table~\ref{tab:probing-vs-neuronpedia}). Appendix~\ref{sec:probing-input-components} reports activation coverage and boundary rejection separately. For a single feature, initialization and average follow-up scans process 0.59M input token positions, 87.4\% fewer than the 4.72M-position corpus pass. Appendix~\ref{sec:probing-volume-accounting} examines processing volume across feature-set sizes. Two examples further show that short-context probing can recover activating tokens absent from available corpus-derived examples and reveal context-dependent activations through adaptive context selection (Appendix~\ref{sec:probing-complementary-evidence}).

\paragraph{Skill accumulation and held-out evaluation.}
\label{sec:skills-self-evolution-main}

We test whether experience from earlier features improves interpretation quality on unseen features and reduces refinement effort. DAFI accumulates skills through sequential interpretation of 120 features, with performance on a fixed held-out suite assessed after every 20 features. Skill accumulation raises the joint pass rate by 34.0 percentage points while reducing refinement rounds (Figure~\ref{fig:skills-self-evolution}). Most of the pass-rate gain appears within the first 20 features, with further accumulation sustaining higher quality at fewer refinement rounds. This suggests that useful strategies emerge early and continue to benefit subsequent interpretations.

\par\medskip\noindent
\begin{minipage}{\linewidth}
\captionsetup{type=figure}
\centering
\begingroup
\renewcommand{\sfdefault}{phv}

\definecolor{skillblue}{HTML}{3A719F}
\definecolor{skillrust}{HTML}{C0784F}
\definecolor{skillink}{HTML}{333537}

\pgfplotsset{
  skillplot/.style={
    width=\linewidth,
    height=0.69\linewidth,
    font=\normalfont\sffamily\fontsize{8}{9}\selectfont,
    label style={
      font=\normalfont\sffamily\bfseries
        \fontsize{8.5}{10}\selectfont,
      text=skillink
    },
    tick label style={
      font=\normalfont\sffamily\fontsize{8}{9}\selectfont,
      text=skillink
    },
    title style={
      at={(0,1.04)},
      anchor=south west,
      inner sep=0pt,
      font=\normalfont\sffamily\bfseries
        \fontsize{9}{10}\selectfont,
      text=skillink
    },
    xlabel={Features interpreted},
    xmin=-3,
    xmax=140,
    xtick={0,20,40,60,80,100,120},
    xticklabels={0,20,40,60,80,100,120},
    axis lines=left,
    axis line style={
      black!60,
      line width=0.45pt,
      -
    },
    tick style={
      black!50,
      line width=0.4pt
    },
    tick align=outside,
    major tick length=2pt,
    xmajorgrids=false,
    ymajorgrids=true,
    grid style={
      black!10,
      line width=0.3pt
    },
    axis background/.style={fill=white},
    clip=false,
  },
}

\tikzset{
  skillvalue/.style={
    font=\normalfont\sffamily\bfseries
      \fontsize{8}{9}\selectfont,
    inner sep=1pt
  },
  skillchange/.style={
    font=\normalfont\sffamily\bfseries
      \fontsize{9}{10}\selectfont,
    fill=white,
    fill opacity=0.94,
    text opacity=1,
    inner sep=2pt
  },
}

\begin{subfigure}[t]{0.47\textwidth}
\centering
\phantomsubcaption
\label{fig:skills-self-evolution-pass-rate}

\begin{tikzpicture}
\begin{axis}[
  skillplot,
  title={(\thesubfigure) Interpretation quality},
  ylabel={Joint pass rate (\%)},
  ymin=50,
  ymax=100,
  ytick={50,60,70,80,90,100},
  yticklabels={50,60,70,80,90,100},
]

\addplot[
  color=skillblue,
  line width=1.1pt,
  mark=*,
  mark size=2pt,
  mark options={
    fill=skillblue,
    draw=white,
    line width=0.35pt
  },
] coordinates {
  (0,58)
  (20,82)
  (40,80)
  (60,84)
  (80,86)
  (100,92)
  (120,92)
};

\node[
  skillvalue,
  text=skillblue,
  anchor=west,
  xshift=5pt,
  yshift=-3pt
] at (axis cs:0,58) {58.0};

\node[
  skillvalue,
  text=skillblue,
  anchor=south east,
  yshift=4pt
] at (axis cs:120,92) {92.0};

\draw[
  {Latex[length=3pt]}-{Latex[length=3pt]},
  color=skillblue,
  line width=0.65pt
] (axis cs:134,58) -- (axis cs:134,92);

\node[
  skillchange,
  text=skillblue,
  anchor=east
] at (axis cs:124,74) {+58.6\%};

\end{axis}
\end{tikzpicture}
\end{subfigure}
\hfill
\begin{subfigure}[t]{0.47\textwidth}
\centering
\phantomsubcaption
\label{fig:skills-self-evolution-rounds}

\begin{tikzpicture}
\begin{axis}[
  skillplot,
  title={(\thesubfigure) Refinement efficiency},
  ylabel={Mean rounds},
  ymin=2.0,
  ymax=4.25,
  ytick={2.0,2.5,3.0,3.5,4.0},
  yticklabels={2.0,2.5,3.0,3.5,4.0},
]

\addplot[
  color=skillrust,
  line width=1.1pt,
  mark=*,
  mark size=2pt,
  mark options={
    fill=skillrust,
    draw=white,
    line width=0.35pt
  },
] coordinates {
  (0,4.00)
  (20,2.60)
  (40,3.20)
  (60,2.70)
  (80,2.64)
  (100,2.70)
  (120,2.42)
};

\node[
  skillvalue,
  text=skillrust,
  anchor=west,
  xshift=5pt,
  yshift=-3pt
] at (axis cs:0,4.00) {4.00};

\node[
  skillvalue,
  text=skillrust,
  anchor=north east,
  yshift=-4pt
] at (axis cs:120,2.42) {2.42};

\draw[
  {Latex[length=3pt]}-{Latex[length=3pt]},
  color=skillrust,
  line width=0.65pt
] (axis cs:134,2.42) -- (axis cs:134,4.00);

\node[
  skillchange,
  text=skillrust,
  anchor=east
] at (axis cs:124,3.32) {\textminus39.5\%};

\end{axis}
\end{tikzpicture}
\end{subfigure}

\endgroup
\caption{Skill accumulation improves held-out interpretation
quality and reduces refinement effort.
(a) Joint pass rate.
(b) Mean agent-loop rounds, including early-stop attempts.
Annotations show relative changes from the initial to
the final evaluation.}
\label{fig:skills-self-evolution}
\end{minipage}
\par\medskip

\paragraph{Skill transfer across model and SAE settings.}
We test whether accumulated skills remain useful in a new model and SAE setting. We evaluate transfer to Qwen3-1.7B-Base with Qwen-Scope SAEs \citep{deng2026qwenscope}, comparing DAFI with and without transferred skills under identical conditions.

Transferred skills improve all three interpretation scores while reducing refinement rounds and token usage (Table~\ref{tab:transfer}), demonstrating their reuse across model and SAE settings. The largest score gain occurs on the output side (7.3 percentage points), while token usage falls by \mbox{14.5\%. Appendix~\ref{sec:transfer-layer-breakdown} details the setup and layer-wise results.}

\par\smallskip\noindent
\begin{minipage}{\linewidth}
\captionsetup{type=table}
\centering
\small
\setlength{\tabcolsep}{3pt}
\renewcommand{\arraystretch}{1.08}
\begin{tabular}{lccccc}
\toprule
\multirow{2}{*}{Condition} & Input & Output & Functional & \multirow{2}{*}{Rounds $\downarrow$} & \multirow{2}{*}{Tokens $\downarrow$} \\
& score (\%) & score (\%) & score & & \\
\midrule
Cold & 75.3 & 41.7 & 3.1 & 1.56 & 434k \\
\rowcolor{oursbluepale}
Skill-seeded & \best{80.5} & \best{49.0} & \best{3.4} & \best{1.45} & \best{371k} \\
\bottomrule
\end{tabular}
\caption{Skill transfer on 100 Qwen-Scope features. Blue marks initialization with transferred skills.}
\label{tab:transfer}
\end{minipage}
\par\medskip

\subsection{What Dual-End Interpretation Reveals about SAE Features}
\label{sec:sae-insights}

We examine how the semantics that activate an SAE feature relate to the output semantics promoted by intervention. Our analysis covers 215 features from the 250-feature pool whose input and output interpretations both pass, allowing us to compare well-supported descriptions of the two endpoints.

\paragraph{Three relations between the endpoints.}
We distinguish three relations (Table~\ref{tab:endpoint-relations}): \textbf{Equivalent} preserves meaning, \textbf{Shift} links distinct meanings through a supported mapping, and \textbf{Break} lacks a supported connection. We classify inspected pairs with LLM assistance and author review.

\par\smallskip\noindent
\begin{minipage}{\linewidth}
\captionsetup{type=table}
\centering
\small
\setlength{\tabcolsep}{5pt}
\renewcommand{\arraystretch}{1.45}
\renewcommand{\tabularxcolumn}[1]{m{#1}}
\begin{tabularx}{\linewidth}{>{\bfseries}m{0.15\linewidth} >{\raggedright\arraybackslash}X}
\toprule
Relation & \textbf{Definition} \\
\midrule
Equivalent & Both endpoints express the same core semantic concept. \\
\rowcolor{oursbluepale}
Shift & Distinct endpoint concepts are linked by an evidence-supported semantic mapping. \\
Break & Both endpoints are supported, but their semantic connection is not established. \\
\bottomrule
\end{tabularx}
\captionsetup{skip=4pt}
\caption{Three semantic relations between well-supported input and output interpretations.}
\label{tab:endpoint-relations}
\end{minipage}

\par\medskip\noindent
\begin{minipage}[t]{0.55\linewidth}
\vspace{0pt}
\textbf{Most endpoint pairs are not equivalent.}
Shift and Break account for \textbf{70.7\%} of the analyzed features (Figure~\ref{fig:taxonomy-results}). Input interpretations alone therefore often miss a feature's output semantics. Dual-end interpretation captures both its activating concept and its output bias.

\smallskip
\textbf{Semantic shifts reveal functional mappings.}
Shift is the largest group at \textbf{63.7\%}. These features respond to one concept while promoting a related but distinct output concept. Functional interpretation explains how this mapping biases generation.
\end{minipage}\hfill
\begin{minipage}[t]{0.42\linewidth}
\vspace{0pt}
\centering
\begingroup
\renewcommand{\sfdefault}{phv}
\definecolor{relshift}{HTML}{315A99}
\definecolor{relshifttop}{HTML}{567FBC}
\definecolor{relshiftbottom}{HTML}{18396D}
\definecolor{relbreaktop}{HTML}{B8C7D9}
\definecolor{relbreakbottom}{HTML}{8DA4BF}
\definecolor{relequivtop}{HTML}{E6EBF2}
\definecolor{relequivbottom}{HTML}{CCD7E5}
\definecolor{relink}{HTML}{303843}
\resizebox{0.99\linewidth}{!}{%
\begin{tikzpicture}[x=1cm,y=1cm,font=\normalfont\sffamily\fontsize{8}{9}\selectfont,text=relink]
  \path[use as bounding box] (-1.83,-1.83) rectangle (3.97,1.95);
  \def\R{1.65}
  \def\r{0.73}
  \pgfmathsetmacro{\shiftend}{90-137/215*360}
  \pgfmathsetmacro{\breakend}{90-152/215*360}
  \foreach \w in {0.5,1,...,7}{
    \path[fill=black,draw=black,opacity=0.008,line width=\w pt,
      even odd rule,transform canvas={xshift=0.5pt,yshift=-1.4pt}]
      (0,0) circle (\R) (0,0) circle (\r);
  }
  \path[top color=relbreaktop,bottom color=relbreakbottom,
    draw=white,line width=0.6pt]
    (\shiftend:\R) arc (\shiftend:\breakend:\R) --
    (\breakend:\r) arc (\breakend:\shiftend:\r) -- cycle;
  \path[top color=relequivtop,bottom color=relequivbottom,
    draw=white,line width=0.6pt]
    (\breakend:\R) arc (\breakend:-270:\R) --
    (-270:\r) arc (-270:\breakend:\r) -- cycle;
  \begin{scope}[shift={(-24.70:0.1815)}]
    \foreach \w in {0.5,1,...,7}{
      \path[fill=black,draw=black,opacity=0.008,line width=\w pt,
        transform canvas={xshift=0.7pt,yshift=-1.8pt}]
        (90:\R) arc (90:\shiftend:\R) --
        (\shiftend:\r) arc (\shiftend:90:\r) -- cycle;
    }
    \path[top color=relshifttop,bottom color=relshiftbottom,
      draw=relshift!20!white,line width=0.35pt]
      (90:\R) arc (90:\shiftend:\R) --
      (\shiftend:\r) arc (\shiftend:90:\r) -- cycle;
    \node[align=center,text=white,font=\sffamily\bfseries\fontsize{7.8}{9}\selectfont]
      at (-24.70:1.20) {63.7\%};
  \end{scope}
  \node[align=center,font=\sffamily\fontsize{7.8}{9}\selectfont]
    at (-151.95:1.15) {7.0\%};
  \node[align=center,font=\sffamily\fontsize{7.8}{9}\selectfont]
    at (-217.26:1.20) {29.3\%};
  \node[align=center,text=relink] at (0,0)
    {\textbf{215}\\features};
  \fill[relshift] (2.13,1.03) circle (0.045);
  \node[anchor=west,text=relshift,font=\sffamily\bfseries\fontsize{9}{10}\selectfont]
    at (2.32,1.03) {Shift};
  \node[anchor=west] at (2.32,0.68) {137 features};
  \fill[relbreakbottom] (2.13,0.10) circle (0.045);
  \node[anchor=west,font=\sffamily\bfseries\fontsize{8}{9}\selectfont]
    at (2.32,0.10) {Break};
  \node[anchor=west] at (2.32,-0.25) {15 features};
  \fill[relequivbottom] (2.13,-0.83) circle (0.045);
  \node[anchor=west,font=\sffamily\bfseries\fontsize{8}{9}\selectfont]
    at (2.32,-0.83) {Equivalent};
  \node[anchor=west] at (2.32,-1.18) {63 features};
\end{tikzpicture}}
\endgroup
\captionsetup{skip=4pt}

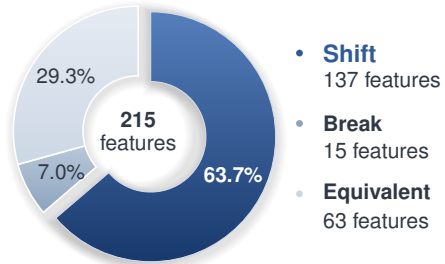
\captionof{figure}{Endpoint relations for 215 features with passing endpoint interpretations.}
\label{fig:taxonomy-results}
\end{minipage}
\par

\label{sec:representative-cases}
\noindent\textbf{Shift (L12 F14100).}
The feature responds to \emph{tourism}, while intervention promotes destinations such as \emph{Malibu}, \emph{Kauai}, and \emph{Puglia}. Its functional interpretation connects tourism to specific destinations.

\noindent\textbf{Equivalent (L24 F7800).}
The feature responds to construction materials, while intervention promotes masonry terms such as \emph{mortar}, \emph{limestone}, and \emph{blocks}. Both endpoints express the same core concept of masonry materials, so the output effect preserves the activating semantics.

\noindent\textbf{Break (L0 F2610).}
The feature responds to \emph{progress} and related forms, while intervention promotes British English forms such as \emph{whilst} and \emph{enquiry}. Both endpoint interpretations are supported, but the evidence does not establish a semantic connection between progress and British English conventions.

\paragraph{A broader view of semantic shifts.}
Table~\ref{tab:shift-cases} compares representative Shift cases with their Neuronpedia explanations. The Neuronpedia explanations shown here either describe activating concepts alone or combine activation and logit evidence in a single semantic description. Through targeted interventions and iterative refinement, DAFI captures the semantics of both activating inputs and promoted outputs more comprehensively and precisely, while explicitly characterizing their functional relation. The resulting functional interpretation clarifies each feature's role during model inference by explaining how it shapes generation in the contexts that activate it.

\par\medskip\noindent
\begin{minipage}{\linewidth}
\centering
\begingroup
\footnotesize
\hypersetup{hidelinks}
\setlength{\tabcolsep}{4pt}
\renewcommand{\arraystretch}{1.18}
\begin{tabularx}{\linewidth}{@{}>{\raggedright\arraybackslash}p{0.08\linewidth}>{\raggedright\arraybackslash}p{0.22\linewidth}>{\columncolor{oursbluepale}\raggedright\arraybackslash}X@{}}
\toprule
\textbf{Feature} & \textbf{Neuronpedia} & \textbf{DAFI Interpretations} \\
\midrule
\href{https://www.neuronpedia.org/gemma-2-2b/24-gemmascope-res-16k/13230}{L24}\newline
\href{https://www.neuronpedia.org/gemma-2-2b/24-gemmascope-res-16k/13230}{F13230}\textsuperscript{*}
& Eligible to qualify
& \textbf{Input $\rightarrow$ Output:} Program eligibility $\rightarrow$ \emph{apply}, \emph{file}. \newline
\textbf{Functional:} Eligibility cues prompt application actions. \\
\href{https://www.neuronpedia.org/gemma-2-2b/12-gemmascope-res-16k/4710}{L12}\newline
\href{https://www.neuronpedia.org/gemma-2-2b/12-gemmascope-res-16k/4710}{F4710}
& Currency and exchange rates
& \textbf{Input $\rightarrow$ Output:} Currency terms $\rightarrow$ \emph{Somalia}, \emph{Iraqi}, \emph{Mexican}. \newline
\textbf{Functional:} Currency cues evoke related countries and regions. \\
\href{https://www.neuronpedia.org/gemma-2-2b/24-gemmascope-res-16k/2430}{L24}\newline
\href{https://www.neuronpedia.org/gemma-2-2b/24-gemmascope-res-16k/2430}{F2430}\textsuperscript{*}
& Game balance and nerfs
& \textbf{Input $\rightarrow$ Output:} Game mechanics $\rightarrow$ \emph{balance}, \emph{balancing}, \emph{nerf}. \newline
\textbf{Functional:} Mechanics cues prompt balancing interventions. \\
\href{https://www.neuronpedia.org/gemma-2-2b/0-gemmascope-res-16k/3420}{L0}\newline
\href{https://www.neuronpedia.org/gemma-2-2b/0-gemmascope-res-16k/3420}{F3420}
& Mswati III, monarch of Swaziland
& \textbf{Input $\rightarrow$ Output:} Mswati and Swaziland $\rightarrow$ \emph{Prince}, \emph{Royal}, \emph{King}. \newline
\textbf{Functional:} A monarch evokes the broader royalty domain. \\
\href{https://www.neuronpedia.org/gemma-2-2b/6-gemmascope-res-16k/15300}{L6}\newline
\href{https://www.neuronpedia.org/gemma-2-2b/6-gemmascope-res-16k/15300}{F15300}
& Trucks and related vehicles
& \textbf{Input $\rightarrow$ Output:} Trucks $\rightarrow$ \emph{load}, \emph{loading}, \emph{freight}, \emph{sleeper}. \newline
\textbf{Functional:} Vehicle cues evoke transport roles and types. \\
\href{https://www.neuronpedia.org/gemma-2-2b/0-gemmascope-res-16k/15510}{L0}\newline
\href{https://www.neuronpedia.org/gemma-2-2b/0-gemmascope-res-16k/15510}{F15510}
& Filenames in various contexts
& \textbf{Input $\rightarrow$ Output:} Filename parameters $\rightarrow$ \emph{open}, \emph{sort}, \emph{docx}, \emph{java}, \emph{adb}. \newline
\textbf{Functional:} Filename cues promote file operations and formats. \\
\href{https://www.neuronpedia.org/gemma-2-2b/0-gemmascope-res-16k/690}{L0}\newline
\href{https://www.neuronpedia.org/gemma-2-2b/0-gemmascope-res-16k/690}{F690}
& Isolation and individualism
& \textbf{Input $\rightarrow$ Output:} \emph{alone} $\rightarrow$ \emph{horrible}, \emph{disgusting}, \emph{cruel}, \emph{terrible}. \newline
\textbf{Functional:} Isolation cues evoke strongly negative affect. \\
\href{https://www.neuronpedia.org/gemma-2-2b/24-gemmascope-res-16k/6930}{L24}\newline
\href{https://www.neuronpedia.org/gemma-2-2b/24-gemmascope-res-16k/6930}{F6930}\textsuperscript{*}
& Legislative codes and statutes
& \textbf{Input $\rightarrow$ Output:} Legal citations $\rightarrow$ \emph{Code}, \emph{amended}, \emph{enacted}, \emph{statutes}. \newline
\textbf{Functional:} Formal citation cues promote legal-domain content. \\
\href{https://www.neuronpedia.org/gemma-2-2b/18-gemmascope-res-16k/3690}{L18}\newline
\href{https://www.neuronpedia.org/gemma-2-2b/18-gemmascope-res-16k/3690}{F3690}\textsuperscript{*}
& ``Get it right''
& \textbf{Input $\rightarrow$ Output:} \emph{get it right} $\rightarrow$ \emph{first}, \emph{primera}, \emph{attempt}. \newline
\textbf{Functional:} Successful performance evokes first-time completion. \\
\bottomrule
\end{tabularx}
\endgroup
\captionof{table}{Representative Shift cases. \textsuperscript{*} marks Neuronpedia's \texttt{np\_acts-logits-general} explanations (activation examples and top-logit evidence); unmarked rows use \texttt{oai\_token-act-pair} explanations (activation examples only). Input, Output, and Functional are DAFI components; L/F denote layer/feature indices, and IDs link to Neuronpedia.}
\label{tab:shift-cases}
\end{minipage}
\par

\subsection{Application Case: Identifying Steerable SAE Features}

We test whether dual-end interpretations help select steerable features by complementing AxBench's concept labels \citep{wu2025axbench} with evidence about intervention-induced output effects.

\par\noindent
\begin{minipage}[t]{0.51\linewidth}
\vspace{0pt}
We compare DAFI-based selection with the output-score filter \citep{arad2025steering} on 500 Concept500 pairs using GemmaScope features from Gemma-2-2B, layer 20. An LLM judge assesses steering viability from DAFI's interpretations and scores. Both filters reuse the same AxBench generations and held-out scores. Coverage is the fraction of pairs retained.

\smallskip
DAFI's filter achieves a \textbf{30.5\%} higher area under the score-coverage curve than the output-score filter and a \textbf{23.2\%} improvement over the unfiltered baseline (Figure~\ref{fig:axbench results}). These gains reflect better feature selection under a fixed steering procedure. Appendix~\ref{sec:output-score-filter-analysis} compares the 2B and 9B protocols.
\end{minipage}\hfill
\begin{minipage}[t]{0.46\linewidth}
\vspace{0pt}
\centering
\includegraphics[width=\linewidth]{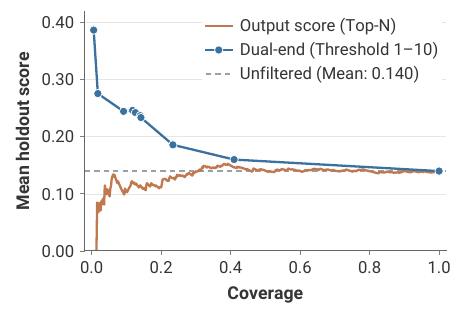}
\captionof{figure}{Concept500 steering scores across coverage. Blue: dual-end. Orange: output-score filter. Grey: unfiltered (0.140).}
\label{fig:axbench results}
\end{minipage}
\par

\section{Conclusion}

DAFI interprets SAE features through input-side, output-side, and functional hypotheses. Equivalent, Shift, and Break relations show that the endpoints often differ, motivating explicit interpretation of their connection. Guided by diagnostic feedback, DAFI coordinates probing and interventions and reuses skills to refine these hypotheses. It improves Input and Output scores over SAGE and Token Change, respectively, while producing functional interpretations. Short-context probing reduces inference volume and uncovers evidence absent from available corpus examples. Reusable skills improve later interpretations and transfer across LLM--SAE settings. Future work will investigate lower-cost interpretation and downstream model editing.

\subsection*{AI use statement}
Generative AI tools were used only for wording refinement. The research conception, methodology, experimental design, analysis, and writing were carried out by the authors. All AI-assisted changes were reviewed and approved by the authors, who take full responsibility for the final content.

\endgroup


\begingroup
\setlength{\emergencystretch}{2em}
\bibliographystyle{iclr2027_conference}
\bibliography{references}
\endgroup
\clearpage
\appendix
\section*{Appendix}

\section{Agent System Details}
\label{sec:agent-details}
The pipeline is decomposed into nine steps. Steps 1 to 4 produce and evaluate the input hypothesis. Steps 5 to 7 run intervention, generate the output hypothesis, and score it. Steps 8 and 9 build and judge the functional hypothesis. The agent supports selective rerun, so a patch that changes output guidance can retain previously collected input evidence rather than rerunning the whole pipeline. This reduces token usage and variability from regenerating unaffected components. Each candidate contains a complete interpretation triplet and its associated scores. Best-round selection retains one candidate as a whole, including the endpoint hypotheses linked by its functional interpretation.

\begin{table}[htbp]
\centering
\footnotesize
\setlength{\tabcolsep}{4pt}
\begin{tabular}{@{}>{\raggedright\arraybackslash}p{0.16\textwidth}>{\raggedright\arraybackslash}p{0.25\textwidth}>{\raggedright\arraybackslash}p{0.32\textwidth}>{\raggedright\arraybackslash}p{0.16\textwidth}@{}}
\toprule
Failure signal & Diagnosis & Candidate repair & Rerun scope \\
\midrule
Low input activation coverage & Positive prompts do not realize the proposed trigger & Narrow the hypothesis to the observed token pattern; rebuild positive prompts & Input design and scoring \\
Low input boundary rejection & The hypothesis also covers adjacent negative contexts & Add explicit exclusions; construct harder boundary prompts & Input design and scoring \\
Low Output score & Output description misses promoted tokens or covers suppressed tokens & Widen delta-token evidence; replace steering prompts; revise output scope & Intervention through functional judgment \\
Low Functional score & Relation is misstated or lacks sufficient evidence & Revise the functional hypothesis if evidence suffices; otherwise adjust context, intervention position or strength, then revise & Functional construction and judgment; intervention onward if new evidence is needed \\
Mixed regression & A repair improves one metric but weakens another & Restore the best prior trace; change repair family or stop & Failed component only \\
\bottomrule
\end{tabular}
\caption{Metric-aware repair policy. The controller changes the smallest component supported by the observed failure and reuses all unaffected evidence.}
\label{tab:repair-policy}
\end{table}

\begin{lstlisting}[language=, caption={Best-round repair and stopping procedure.}, label={lst:agent-loop}]
state = run_initial_pipeline(feature)
best = state
for round in 1..10:
    failed = metrics_below_threshold(state)
    if failed is empty: break
    patch = propose_smallest_supported_repair(state, failed)
    candidate = selective_rerun(state, patch)
    best = rank_by_joint_then_metrics(best, candidate)
    if no_metric_progress(candidate):
        change_repair_family_or_stop()
    state = candidate
return best
\end{lstlisting}

\subsection{Claude Code Baseline Instructions}
\label{sec:claude-code-instructions}

Listing~\ref{lst:claude-code-instructions} reproduces the Claude Code baseline's complete \texttt{CLAUDE.md}, specifying its interpretation target, pipeline, quality gates, iteration protocol, and logging requirements.

\begin{lstlisting}[
  language=,
  basicstyle=\tiny\ttfamily,
  caption={Complete task instructions provided to the Claude Code baseline through \texttt{CLAUDE.md}.},
  label={lst:claude-code-instructions}
]
# SAE Feature Interpretation -- Agent Task

You are interpreting a single Sparse Autoencoder (SAE) feature of the Gemma-2-2b
language model. A 9-step pipeline is already set up in this directory. Running it
produces an interpretation in three parts and writes a `trace.json`:

- **Input side** -- hypotheses about *what kind of text makes the feature activate*,
  scored by how reliably purpose-built sentences actually activate it.
- **Output side** -- hypotheses about *what the feature promotes* when its activation
  is steered up, scored by how well the predicted tokens match the real token shifts.
- **Functional interpretation** -- a short causal explanation linking the input cause to the
  output effect, judged on a 1--5 scale.

## Running the pipeline

From this directory:

```bash
./run_pipeline.sh <LAYER> <FEATURE> <TIMESTAMP>
```

runs all 9 steps once and writes everything under
`logs/layer-<LAYER>/feature-<FEATURE>/<TIMESTAMP>/`, including `trace.json`.

Each step is also an independent script (`step1_*.py` ... `step9_*.py`). Run
`python <step>.py --help` to see what options it accepts. Steps share the timestamp
directory: a later step reads the outputs that earlier steps wrote there. Re-running a
step overwrites only that step's output, so after changing one step you must re-run it
and every step after it (step 9 rewrites `trace.json`).

The LLM-based steps (2, 3, 6, 7, 8, 9) and the GPU-based steps (4, 5) are already
configured -- you do not need to set model paths, API keys, or the SAE service.

**Pipeline runtime (important for how you call it):** each step takes up to ~1--2
minutes, and a full `run_pipeline.sh` pass takes roughly 8--10 minutes. When you run
`run_pipeline.sh`, give the command a long timeout (about 600000 ms / 10 minutes) so it
is not killed early. If you prefer, run the steps one at a time instead -- each
individual step finishes well within a couple of minutes.

## Your goal

Produce an interpretation in which **all three quality gates pass**:

| Gate | Meaning | Pass condition |
|------|---------|----------------|
| Gate 1 (input)  | activation rate of the best input hypothesis | >= 0.8 |
| Gate 1 (input)  | boundary non-activation rate of that hypothesis | >= 0.8 |
| Gate 2 (output) | Output score of the functional interpretation's best endpoint pair | >= 0.5 |
| Gate 3 (functional)  | Functional score of the best endpoint pair | >= 4 (of 5) |

Read the gate values with the provided helper (this is the authoritative scorer --
use it, do not hand-roll your own jq):

```bash
python show_gates.py logs/layer-<LAYER>/feature-<FEATURE>/<TIMESTAMP>/trace.json
```

It prints each gate's value, whether it passes, and `all_gates_passed`.

## The iteration loop -- at most 10 rounds

1. **Round 1**: run the full pipeline once with a base timestamp (e.g. `r1`).
2. **Diagnose**: read `trace.json`. Decide which gate(s) fail and form your own
   hypothesis about *why*. The per-round evidence lives under the timestamp
   directory (the `input/`, `output/`, `chain/` subfolders hold per-sentence
   activations, the steered-token lists, and the functional-interpretation judge's reasoning) -- inspect
   whatever you need to support your diagnosis.
3. **Adjust and re-run as a NEW round**: decide a change that might fix the failing
   gate, then run another round under a **new timestamp** suffixed `_r2`, `_r3`, ...
   You may either re-run the whole pipeline, or copy the previous round's directory
   to the new timestamp and re-run only the step(s) you changed plus everything
   downstream of them.
4. Repeat until **all gates pass** or you have completed **round 10**, then stop.

**Round-logging rule (important):** every pipeline execution must live in its own
round directory (`r1`, then `_r2`, `_r3`, ...). Never re-run steps repeatedly inside one
round directory, and never run the pipeline without recording it as a numbered round.
One round = one timestamp directory.

How to find out *what* you can change: read each step's `--help` and, if useful, its
source. Deciding which knobs to turn for which failure is your job -- there is no fixed
recipe.

## Keep an experience log

Maintain `EXPERIENCE.md` in this directory. After each round, append: the round
number, the gate numbers you observed, what you changed and why, and what you concluded.
Use it so you do not repeat attempts that already failed.

## When you finish

Print a final JSON object summarising the run:

```json
{"layer": "<LAYER>", "feature": "<FEATURE>", "rounds_run": <N>,
 "final_gates": {"g1_act": ..., "g1_bnd": ..., "g2": ..., "g3": ...},
 "all_gates_passed": true|false}
```
\end{lstlisting}

\section{Refinement Dynamics and Stability}
\label{sec:refinement-dynamics}

\begin{figure}[H]
\centering
\begin{tikzpicture}[font=\sffamily]
\begin{axis}[
  width=0.76\linewidth,
  height=0.48\linewidth,
  ybar,
  bar width=8pt,
  xlabel={Executed refinement rounds},
  ylabel={Features},
  xmin=-0.5, xmax=10.5,
  ymin=0,
  xtick={0,1,2,3,4,5,10},
  axis y line*=left,
  axis x line*=bottom,
  nodes near coords,
  every node near coord/.append style={font=\scriptsize},
  tick label style={font=\small},
  label style={font=\small},
]
\addplot[draw=oursblue, fill=oursblue!65] coordinates {(0,20) (1,40) (2,85) (3,29) (4,30) (5,15) (6,10) (7,8) (8,2) (9,3) (10,8)};
\end{axis}
\begin{axis}[
  width=0.76\linewidth,
  height=0.48\linewidth,
  xmin=-0.5, xmax=10.5,
  ymin=0, ymax=100,
  axis y line*=right,
  axis x line=none,
  xtick=\empty,
  ylabel={Cumulative share (\%)},
  ytick={0,20,40,60,80,100},
  tick label style={font=\small},
  label style={font=\small},
]
\addplot[shiftorange, line width=1.1pt, mark=*] coordinates {(0,8.000) (1,24.000) (2,58.000) (3,69.600) (4,81.600) (5,87.600) (6,91.600) (7,94.800) (8,95.600) (9,96.800) (10,100.000)};
\end{axis}
\end{tikzpicture}
\caption{Distribution and cumulative share of executed refinement rounds over 250 features. Bars show feature counts, and the line shows the cumulative share.}
\label{fig:refinement-rounds-appendix}
\end{figure}
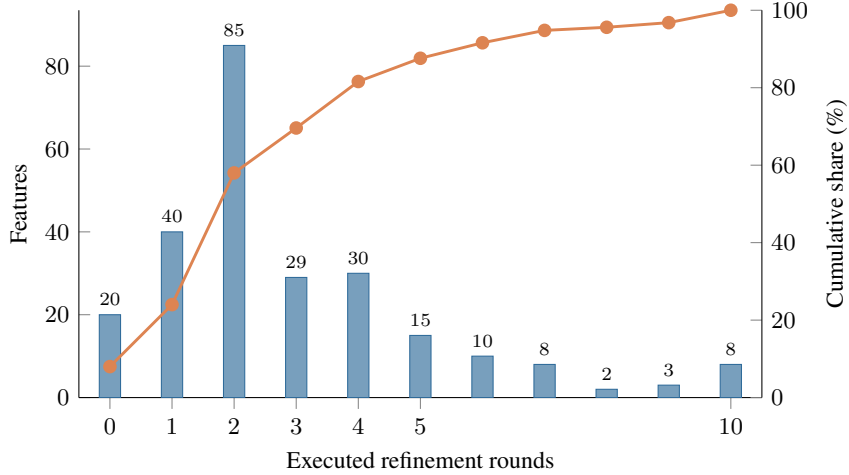

Figure~\ref{fig:refinement-rounds-appendix} reports the distribution and cumulative share of executed refinement rounds. Across 250 features, the agent enters 2.92 rounds per feature on average; 145 features record no more than two rounds, whereas 8 reach the 10-round cap. Most runs finish early, with a long tail of difficult cases.

Table~\ref{tab:score-stability-250} reports per-feature changes from the initial scores to those of the selected adaptive interpretations. The minimum of the two input components decreases for only 0.4\% of features, while Output score and Functional score decrease for 5.2\% and 6.4\%, respectively. Together with the aggregate improvements in all three interpretation components, these results indicate that refinement is generally stable, although individual diagnostic scores are not guaranteed to improve monotonically.

\begin{table}[htbp]
\centering
\footnotesize
\setlength{\tabcolsep}{7pt}
\begin{tabular}{@{}lrrr@{}}
\toprule
Metric & Improved, $n$ (\%) & Unchanged, $n$ (\%) & Decreased, $n$ (\%) \\
\midrule
Minimum input component & 88 (35.2) & 161 (64.4) & 1 (0.4) \\
Output score & 174 (69.6) & 63 (25.2) & 13 (5.2) \\
Functional score & 138 (55.2) & 96 (38.4) & 16 (6.4) \\
\bottomrule
\end{tabular}
\caption{Diagnostic changes from initial to selected adaptive interpretations for 250 features. The minimum input component is the smaller of activation coverage and boundary rejection.}
\label{tab:score-stability-250}
\end{table}

\section{Explanation and Judge Templates}
\label{sec:explanation-and-judge-templates}
Each generation call is constrained to one component of the explanation object. The input prompt receives activating tokens and contexts and asks for a concise hypothesis with explicit exclusions. The output prompt receives positive and negative intervention deltas and asks for the narrowest description that covers promoted evidence without also covering suppressed evidence. The functional-interpretation prompt receives both hypotheses and their supporting evidence, but it is not allowed to introduce an unobserved intermediate mechanism.

\begin{table}[t]
\centering
\small
\setlength{\tabcolsep}{4pt}
\begin{tabular}{@{}cp{0.82\linewidth}@{}}
\toprule
Score & Criterion \\
\midrule
1 & No supported relation between the input and output hypotheses. \\
2 & The proposed relation is contradicted by the available evidence. \\
3 & The relation is plausible but only weakly grounded in the evidence. \\
4 & A clear relation is supported by representative evidence from both endpoints. \\
5 & The relation is strongly supported, specific, and free of salient contradictions. \\
\bottomrule
\end{tabular}
\caption{Five-point rubric used by the functional judge.}
\label{tab:m3-rubric}
\end{table}

For each feature, the judge returns a Functional score, a brief evidence-based rationale, and a relation label. These outputs are used for automated diagnosis but are withheld from the human annotators, ensuring that the human ratings are collected independently of the LLM judgment.

\subsection{Semantic Relation Classification}
\label{sec:relation-classification-prompt}

For the semantic-relation analysis in Section~\ref{sec:sae-insights}, we apply the classifier only to features whose input-side and output-side components both pass. For each feature, the classifier receives the selected input-side interpretation, output-side interpretation, functional interpretation, and functional \mbox{judgment. The classification prompt is shown in Listing~\ref{lst:relation-classification-prompt}.}

\begin{lstlisting}[
  language=,
  caption={Prompt for classifying the semantic relation between the input and output endpoints.},
  label={lst:relation-classification-prompt}
]
You are classifying the semantic relation between the input and output endpoints of one SAE feature.

Input-side interpretation:
{input_interpretation}

Output-side interpretation:
{output_interpretation}

Functional interpretation:
{chain_interpretation}

Functional score:
{m3_score}

Functional rationale:
{m3_rationale}

Choose exactly one of the following relation types.

Equivalent:
The two endpoints preserve the same underlying concept or semantic function. The surface terms may differ, but the output effect directly expresses, completes, or amplifies the concept detected on the input side. Use this label for near semantic identity or direct structural continuation. A broad topical association alone is not sufficient.

Shift:
The two endpoints express different concepts or semantic roles, but there is an identifiable semantic, contextual, or functional mapping from the activating context to the promoted output behavior.

Break:
Both endpoint interpretations are individually supported, but the available evidence does not establish a coherent semantic connection between the activating concept and the promoted output semantics.

Classify the relation by comparing the meanings at the two endpoints and the evidence supporting their connection. Use Break when no coherent semantic relation is supported, Equivalent when the core meaning is preserved, and Shift when distinct meanings are linked by an interpretable semantic mapping.

Apply the following decision order:
1. Choose Break if the available evidence does not support a coherent semantic connection between the two endpoints.
2. Otherwise, choose Equivalent if the same underlying concept or semantic function is preserved.
3. Otherwise, choose Shift.

Return JSON only:
{"relation_type": "<Equivalent|Shift|Break>"}
\end{lstlisting}

\section{Layer-wise Refinement Analysis}
\label{sec:layer-level-notes}
We examine refinement across five layers from shallow to deep using 250 GemmaScope features. Refinement improves every reported metric at all five layers (Table~\ref{tab:appendix-layer-summary-250}). Mean Input score rises from 83.3\% to 93.3\%, mean Output score from 32.5\% to 69.5\%, and mean Functional score from 3.0 to 4.0. The joint pass count increases from 20 of 250 (8.0\%) to 200 of 250 (80.0\%). Final Input score exceeds 90\% at every layer, while final mean Output score varies only from 66.4\% to 71.9\%, indicating consistent performance across depth. Table~\ref{tab:appendix-layer-summary-250} reports the two Input score components separately. Appendices~\ref{sec:threshold-robustness} and~\ref{sec:refinement-dynamics} detail threshold robustness, refinement rounds, and metric stability.

\begin{table}[htbp]
\centering
\scriptsize
\setlength{\tabcolsep}{1.5pt}
\begin{tabular}{@{}lcccccccccc@{}}
\toprule
& \multicolumn{2}{c}{Activation coverage} & \multicolumn{2}{c}{Boundary rejection} & \multicolumn{2}{c}{Output score} & \multicolumn{2}{c}{Functional score} & \multicolumn{2}{c}{Joint pass (\%)} \\
\cmidrule(lr){2-3}\cmidrule(lr){4-5}\cmidrule(lr){6-7}\cmidrule(lr){8-9}\cmidrule(lr){10-11}
Layer & Init. & Adapt. & Init. & Adapt. & Init. & Adapt. & Init. & Adapt. & Init. & Adapt. \\
\midrule
L0 & 0.920 & 0.960 & 0.936 & 0.960 & 0.230 & 0.682 & 2.920 & 3.940 & 10.0 & 84.0 \\
L6 & 0.844 & 0.952 & 0.788 & 0.920 & 0.313 & 0.714 & 2.720 & 3.860 & 6.0 & 74.0 \\
L12 & 0.840 & 0.948 & 0.804 & 0.900 & 0.276 & 0.664 & 2.940 & 3.920 & 6.0 & 74.0 \\
L18 & 0.864 & 0.904 & 0.748 & 0.908 & 0.338 & 0.719 & 2.620 & 3.940 & 4.0 & 78.0 \\
L24 & 0.864 & 0.940 & 0.736 & 0.940 & 0.468 & 0.697 & 3.680 & 4.280 & 14.0 & 90.0 \\
\midrule
All & 0.866 & 0.941 & 0.802 & 0.926 & 0.325 & 0.695 & 3.000 & 4.000 & 8.0 & 80.0 \\
\bottomrule
\end{tabular}
\caption{Layer-wise Input-score components, scores, and joint pass rates for 250 features.}
\label{tab:appendix-layer-summary-250}
\end{table}

\subsection{Layer Depth Separates Input-Side Reliability from Functional Interpretation Quality}
Layer depth changes which component of a feature explanation is easiest to recover. As shown in Table~\ref{tab:appendix-layer-summary-250}, Layer~0 achieves the highest refined activation coverage and boundary rejection, both 0.960, indicating that its input-side explanations align most reliably with feature activations under our evaluation. However, accurate input-side explanations do not necessarily imply a clearer relationship between feature activations and intervention-induced output effects. The final Functional score is 3.94 at both Layer~0 and Layer~18, compared with 4.28 at Layer~24.

Before refinement, Layer~24 also has a higher Output score than Layer~0, 0.468 versus 0.230. After refinement, their scores become similar, at 0.697 and 0.682, suggesting that targeted refinement can reduce layer-wise disparities in output-side interpretation quality. Layer~18 shows the largest Functional score improvement, from 2.62 to 3.94, while Layer~24 achieves both the highest final Functional score and the highest joint pass rate. These results distinguish input-side reliability from the clarity of the input--output relationship: strong activation-side recoverability does not by itself imply the clearest functional relationship to downstream effects. This is a layer-level tendency in explanation recoverability, not evidence that every deep feature is a better intervention target.

Qualitative inspection suggests that Layer~0 features often capture subword, casing, punctuation, or local morphology, while Layer~6 features often capture lexical associations. Layer~12 behaves as a transition layer in which causal narratives can be coherent but output token effects remain indirect. Layer~18 contains many deep semantic features whose explanations improve substantially after several agent rounds. Layer~24 is closest to the output side and reaches the highest joint pass rate.

\section{Skill Accumulation Details}
\label{sec:skills-self-evolution}
This section supplements the held-out skill-accumulation results in Section~\ref{sec:skills-self-evolution-main}. We accumulate skills over 120 features and evaluate each checkpoint on a fixed held-out set of 50 features, with ten features per layer. A feature passes only if the input-side, output-side, and functional-interpretation gates all pass. The evaluation set and scoring configuration remain fixed across checkpoints.

\begin{table}[htbp]
\centering
\footnotesize
\setlength{\tabcolsep}{4.5pt}
\begin{tabular}{@{}lrrrrrrr@{}}
\toprule
Features interpreted & 0 & 20 & 40 & 60 & 80 & 100 & 120 \\
\midrule
Joint pass (\%) & 58.0 & 82.0 & 80.0 & 84.0 & 86.0 & 92.0 & 92.0 \\
Mean agent-loop rounds & 4.00 & 2.60 & 3.20 & 2.70 & 2.64 & 2.70 & 2.42 \\
\bottomrule
\end{tabular}
\caption{Held-out joint pass rate and mean refinement rounds during skill accumulation.}
\label{tab:skills-self-evolution}
\end{table}

The joint pass rate rises from 58.0\% to 92.0\% as experience accumulates, a gain of 34.0 percentage points. Over the same checkpoints, the mean number of agent-loop rounds entered per feature decreases overall from 4.00 to 2.42, a 39.5\% reduction. Despite intermediate fluctuations, the endpoint comparison shows improved interpretations with less agent effort.

\section{Short-Context Probing Details}

\subsection{Input-Score Component Breakdown}
\label{sec:probing-input-components}

Table~\ref{tab:probing-input-components} separates the reported Input scores into activation coverage and boundary rejection.

\begin{table}[H]
\centering
\footnotesize
\setlength{\tabcolsep}{3.5pt}
\begin{tabular}{@{}lcccc@{}}
\toprule
\multirow{2}{*}{Initialization} & \multicolumn{2}{c}{Activation coverage (\%)} & \multicolumn{2}{c}{Boundary rejection (\%)} \\
\cmidrule(lr){2-3}\cmidrule(lr){4-5}
& Initial & Final & Initial & Final \\
\midrule
Neuronpedia & 83.8 & 91.6 & 80.4 & 91.2 \\
Short-context probing & 68.2 & 84.8 & 77.8 & 92.6 \\
\bottomrule
\end{tabular}
\caption{Input-score component breakdown for Neuronpedia and one-template short-context-probing initialization on 100 matched GemmaScope features.}
\label{tab:probing-input-components}
\end{table}

\subsection{Evidence-Collection Volume}
\label{sec:probing-volume-accounting}

We measure evidence-collection volume in \emph{input token positions during LLM--SAE inference}: the sum of the tokenized lengths of the sequences processed to collect activation evidence. This measure is not FLOPs or wall-clock cost and excludes tokens used in LLM API calls by the interpretation agent. A single model pass can expose the required hidden states for multiple SAE layers and target features, so we do not multiply a shared corpus or probing scan by the number of evaluated layers or features.

For short-context probing, we count the unique scans in the finalized 100-feature trajectories. The initial full-vocabulary scan can be shared across all target features. Follow-up scans, however, were requested within concurrently running per-feature interpretation processes and were executed independently; we therefore count each unique successful follow-up tool call recorded in the trajectories.

For the shared initialization, Gemma-2-2B has a vocabulary of 256,000 tokens. Excluding six special-token IDs leaves 255,994 candidate tokens. Each initial sequence contains the \verb|<bos>| token followed by one candidate token, giving
\begin{equation}
N_{\mathrm{init}}=255{,}994,
\qquad
C_{\mathrm{init}}=2N_{\mathrm{init}}=511{,}988.
\end{equation}
For a follow-up scan $s$, we compute
\begin{equation}
C_s=N_s\left(L^{\mathrm{pre}}_s+1+L^{\mathrm{suf}}_s\right),
\end{equation}
where $N_s$ is the number of candidate-token sequences and $L^{\mathrm{pre}}_s$ and $L^{\mathrm{suf}}_s$ are the tokenized prefix and suffix lengths around the candidate token. The five features that triggered follow-up scans were Layer~18 features 2700, 3450, 7740, 10320, and 12150; the other 95 features triggered none. Table~\ref{tab:probing-volume-audit} reports the resulting aggregation. All 29 follow-up scans had distinct, available output artifacts.

\begin{table}[H]
\centering
\small
\setlength{\tabcolsep}{5pt}
\begin{tabular}{lrrr}
\toprule
Component & Scans & Sequences & Input token positions \\
\midrule
Shared initialization & 1 & 255,994 & 511,988 \\
Follow-up: full vocabulary & 5 & 1,279,970 & 4,863,886 \\
Follow-up: random sample & 18 & 855,000 & 3,410,000 \\
Follow-up: manual candidates & 6 & 71 & 270 \\
\cmidrule(lr){1-4}
Follow-up subtotal & 29 & 2,135,041 & 8,274,156 \\
\textbf{Short-context total} & \textbf{30} & \textbf{2,391,035} & \textbf{8,786,144} \\
\bottomrule
\end{tabular}
\caption{Input volume for short-context probing in the finalized 100-feature experiment. Initialization is shared across targets; follow-up scans count unique successful calls in the final trajectories.}
\label{tab:probing-volume-audit}
\end{table}

For the Neuronpedia comparison, the Gemma-2-2B GemmaScope residual-stream dashboard reports 36,864 prompts of 128 tokens from \texttt{monology/pile-uncopyrighted} for each evaluated SAE layer.\footnote{Neuronpedia reports this configuration on the corresponding feature dashboard pages; for example, \url{https://www.neuronpedia.org/gemma-2-2b/18-gemmascope-res-16k/1118} (accessed August 10, 2026). We verified the same dashboard configuration for layers 0, 6, 12, 18, and 24.} Because the same corpus pass can provide hidden states for all five evaluated layers, this gives
\begin{equation}
C_{\mathrm{NP}}=36{,}864\times128=4{,}718{,}592
\end{equation}
input token positions. To characterize targeted use, we amortize the observed follow-up volume across the 100 evaluated features. The average volume for interpreting one target feature is
\begin{equation}
C_{\mathrm{target}}
=511{,}988+\frac{8{,}274{,}156}{100}
=594{,}730,
\qquad
1-\frac{594{,}730}{4{,}718{,}592}=87.4\%.
\end{equation}
Under this average-case accounting, short-context probing uses 87.4\% fewer input token positions for one targeted feature, making it suitable for small feature sets on demand.

The scaling relation reverses for high-throughput interpretation. Across all 100 targets, the shared initialization and all observed follow-up scans process
\begin{equation}
C_{100}=511{,}988+8{,}274{,}156=8{,}786{,}144
\end{equation}
input token positions, which is 86.2\% more than the 4.72M-position Neuronpedia corpus pass. The current experiment runs feature interpretations concurrently but records each feature's adaptively requested scan only for that feature. A shared inference service could instead record every requested scan for all active target features, amortizing adaptive probing across concurrent interpretations. We leave this infrastructure optimization to future work.

\subsection{Complementary Evidence from Short-Context Probing}
\label{sec:probing-complementary-evidence}

Short-context probing and corpus-based initialization provide different forms of activation evidence. Corpus-based evidence is limited to the contexts represented in the available activation examples, whereas short-context probing directly tests candidate tokens across controlled templates. The following cases illustrate adaptive context discovery and evidence complementarity.

\paragraph{Layer 18, Feature 1830.}
This feature illustrates how controlled vocabulary coverage can expose activation evidence missing from available corpus-derived examples. The 45 Neuronpedia activation examples were dominated by the English token \verb|the|, and none contained any of the 22 tokens activated by the initial full-vocabulary \verb|<bos>{token}| scan. Representative activations include \verb|InThe| (40.25), \verb|inthe| (29.00), \verb|Nei| (21.75), \verb|Nella| (17.75), and \verb|Nel| (15.25). Together with related Italian, French, Arabic, Thai, and Hebrew forms, these tokens revealed a multilingual locative pattern not represented in the available Neuronpedia examples. The most specific hypothesis---Italian prepositional articles such as \emph{nel}, \emph{nella}, \emph{nei}, and \emph{nelle}---achieved activation coverage of 1.0 and boundary rejection of 1.0. These tokens may occur in the underlying corpus, but they \mbox{are absent from the evidence available to the interpreter.}

\paragraph{Layer 18, Feature 3450.}
This feature provides a direct example of adaptive context discovery within the 100-feature experiment. The initial full-vocabulary scan with \verb|<bos>{token}| evaluated 255,994 candidate tokens but found no activation. After three consecutive Gate~1 failures, the agent tested several context prefixes, including \verb|After|, \verb|Although|, \verb|Despite|, and \verb|In spite|. A 50,000-token random scan with \verb|<bos>Despite {token}| first found four activating tokens; expanding this template to the full vocabulary \mbox{found 39 activating tokens and a peak activation of 9.75.}

\begin{table}[H]
\centering
\footnotesize
\setlength{\tabcolsep}{4pt}
\begin{tabular}{lrrr}
\toprule
Template & Candidates & Active & Max. act. \\
\midrule
\verb|<bos>{token}| & 255,994 & 0 & 0.00 \\
\verb|<bos>Despite {token}| (random) & 50,000 & 4 & 7.75 \\
\verb|<bos>Despite {token}| (full) & 255,994 & 39 & 9.75 \\
\bottomrule
\end{tabular}
\caption{Context discovery for Layer~18, Feature~3450. Both full scans cover the same vocabulary.}
\label{tab:l18-f3450-context-discovery}
\end{table}

Representative full-scan activations include \verb|<bos>Despite soundness| (9.75), \verb|<bos>Despite flawless| (9.06), \verb|<bos>Despite manageable| (9.00), and \verb|<bos>Despite reassurance| (6.75). Because the initial scan over the same vocabulary had a maximum activation of zero, these observations isolate the effect of adding the \verb|Despite| context. The trajectory selected this full-vocabulary scan as its input observation, and the final input-side hypothesis achieved activation coverage of 1.0 and boundary rejection of 1.0. This case demonstrates that the agent can adapt its probing context after an initially uninformative scan. It directly supports context-dependent activation discovery; the semantic interpretation of the discovered pattern remains a hypothesis evaluated by the subsequent pipeline.

\paragraph{Layer 6, Feature 8800.}
This case comes from an earlier six-template diagnostic run and is not part of the 100-feature volume accounting above. The available Neuronpedia activation examples for this feature do not include the token \texttt{endforeach}. This absence does not establish that the token never occurs in the underlying corpus; it only shows that the token is not represented in the available activation examples. By contrast, the initial six-template short-context scan identifies \texttt{endforeach} as the strongest aggregate token for this feature. It appears among the top two tokens for all six templates, ranking first for five templates and second for the remaining template. Its activation ranges from 23.375 to 32.500, with a mean of 27.458 and a maximum of 32.500. Short-context probing thus recovers repeatable activation evidence absent from the available corpus examples.

\section{Human Validation Details}
\label{sec:human-validation}
We evaluate functional-interpretation plausibility on a stratified 40-feature subset spanning layers, final metric states, and low, middle, and high Functional scores. Two researchers with experience in SAE and interpretability research annotate the same examples independently. Both are blinded to the system scores and to each other's labels. Each annotator rates input fit, output fit, and functional-interpretation plausibility on a five-point scale, and assigns one relation label among Equivalent, Shift and Break. We define a human plausibility pass as a plausibility rating of at least 4. For ordinal ratings, we report Spearman correlation, agreement within one point, and quadratic-weighted Cohen's $\kappa$. For relation labels and binary pass decisions, we report unweighted Cohen's $\kappa$.

For each annotator, we compute Spearman correlation across the same 40 features, assigning average ranks to tied scores. The system score for each feature is the Functional score associated with the selected complete interpretation triplet in its final trace. The input-side, output-side, and functional interpretations are retained together, and the human annotators assess this same selected triplet. We compare this shared system-score vector separately with each annotator's plausibility ratings, without averaging or pooling the human ratings. Binary agreement is also computed separately by applying the score-at-least-4 threshold to both the system score and the human plausibility rating, yielding 38/40 agreements for annotator 1 and 37/40 for annotator 2.

The annotators agree strongly on functional-interpretation plausibility, with Spearman correlation 0.865, quadratic-weighted $\kappa$ 0.841, and 92.5\% agreement within one point. Their binary pass decisions also agree on 92.5\% of features, with Cohen's $\kappa$ 0.826. Relation-label agreement is lower but remains substantial at 70.0\%, with $\kappa$ 0.576. The system Functional score correlates with both annotators' plausibility ratings and matches their binary decisions on more than 92\% of features. We therefore interpret this study as evidence of agreement between feature-level Functional score and human judgments of functional-interpretation plausibility.

\begin{table}[htbp]
\centering
\footnotesize
\begin{tabular}{@{}lc@{}}
\toprule
Statistic & Value \\
\midrule
Inter-annotator plausibility Spearman & 0.865 \\
Inter-annotator plausibility quadratic-weighted $\kappa$ & 0.841 \\
Plausibility-rating agreement within one point (\%) & 92.5 \\
Inter-annotator pass agreement (\%) & 92.5 \\
Inter-annotator pass Cohen's $\kappa$ & 0.826 \\
Relation-label agreement (\%) & 70.0 \\
Relation-label Cohen's $\kappa$ & 0.576 \\
System-human Spearman, annotators 1 / 2 & 0.809 / 0.777 \\
System-human pass agreement (\%), annotators 1 / 2 & 95.0 / 92.5 \\
\bottomrule
\end{tabular}
\caption{Independent human validation on 40 stratified features. Both system comparisons use the same feature-level Functional scores and are reported separately for the two annotators. Agreement within one rating point and pass agreement compare the annotators; ratings of at least 4 pass.}
\label{tab:human-validation}
\end{table}

\section{Cross-Model Judge Consistency}
\label{sec:cross-model-judge-consistency}

Because the Output and Functional scores contain LLM-judged components, we test whether their scores are robust to the choice of judge. We re-evaluate all eligible output-side and functional hypotheses in the final traces of the same 250 GemmaScope features with GPT-5.6-Sol, using the same evidence and scoring rules as the original DeepSeek-V4-Pro evaluation. The original scores and rationales are withheld from GPT-5.6-Sol. We report Pearson and Spearman correlations, mean absolute error (MAE), and pass agreement at the fixed Output and Functional thresholds of 0.5 and 4.

\begin{table}[htbp]
\centering
\footnotesize
\setlength{\tabcolsep}{6pt}
\begin{tabular}{@{}lcc@{}}
\toprule
Metric & Output score & Functional score \\
\midrule
Pearson correlation & 0.783 & 0.710 \\
Spearman correlation & 0.811 & 0.649 \\
Mean absolute error & 0.129 & 0.531 \\
Pass-decision agreement (\%) & 87.1 & 80.4 \\
\bottomrule
\end{tabular}
\caption{Cross-model agreement between GPT-5.6-Sol and the original DeepSeek-V4-Pro judgments on the final traces of 250 GemmaScope features. Pass-decision agreement uses the fixed thresholds of 0.5 for Output score and 4 for Functional score.}
\label{tab:cross-model-judge-consistency}
\end{table}

GPT-5.6-Sol agrees with DeepSeek-V4-Pro on 87.1\% of Output score and 80.4\% of Functional score pass decisions. The lower rank correlation for Functional score indicates that this metric is more sensitive to the judge model than Output score. Nevertheless, the results provide evidence that the thresholded conclusions are reasonably stable across the two judges. This tests robustness to judge choice; Appendix~\ref{sec:human-validation} separately examines agreement with human judgments.

\section{Cross-SAE Layer Breakdown}
\label{sec:transfer-layer-breakdown}
Table~\ref{tab:transfer-layer} reports the layer-level transfer results on Qwen3-1.7B-Base with Qwen-Scope W32K residual-stream SAEs using Top-$k$ activation with $k=50$. The results show quality and efficiency gains from transferred skills, with variation across layers.

\begin{table}[htbp]
\centering
\scriptsize
\setlength{\tabcolsep}{3pt}
\begin{tabular}{@{}lccccccc@{}}
\toprule
Condition & Layer & $n$ & \shortstack{Input score\\Activation} & \shortstack{Input score\\Boundary} & \shortstack{Output\\score} & \shortstack{Functional\\score} & \shortstack{Mean\\rounds} \\
\midrule
Cold & 0  & 20 & 0.78 & 0.87 & 0.446 & 3.20 & 1.40 \\
Cold & 7  & 20 & 0.54 & 0.86 & 0.176 & 3.15 & 1.50 \\
Cold & 14 & 20 & 0.68 & 0.71 & 0.518 & 3.10 & 1.30 \\
Cold & 21 & 20 & 0.62 & 0.81 & 0.371 & 2.80 & 1.75 \\
Cold & 27 & 20 & 0.90 & 0.80 & 0.573 & 3.35 & 1.85 \\
Cold & All & 100 & 0.704 & 0.810 & 0.417 & 3.12 & 1.56 \\
\midrule
Skill-seeded & 0  & 20 & 0.88 & 0.89 & 0.475 & 3.30 & 1.50 \\
Skill-seeded & 7  & 20 & 0.57 & 0.89 & 0.327 & 3.20 & 1.35 \\
Skill-seeded & 14 & 20 & 0.70 & 0.72 & 0.535 & 3.25 & 1.20 \\
Skill-seeded & 21 & 20 & 0.69 & 0.90 & 0.506 & 3.55 & 1.55 \\
Skill-seeded & 27 & 20 & 0.95 & 0.89 & 0.605 & 3.60 & 1.65 \\
Skill-seeded & All & 100 & 0.758 & 0.858 & 0.490 & 3.38 & 1.45 \\
\bottomrule
\end{tabular}
\caption{Input-score components and layer-level transfer results on Qwen-Scope. Each layer contains 20 sampled features; All gives the 100-feature aggregate.}
\label{tab:transfer-layer}
\end{table}

\section{Budget Comparison Details}
\label{sec:budget-details}
Claude Code uses DeepSeek-V4-Pro and is evaluated on 25 features, with five features sampled from each layer. In the restricted setting, DAFI's per-feature token usage sets Claude Code's round-start budget. Claude Code always completes one round, even if it exceeds the budget, and cannot start another after reaching the budget. Under this restriction, Claude \mbox{Code uses 77.06M tokens in total and passes 2 features.}

Table~\ref{tab:baseline-input-components} separates the main baseline Input scores into their two components.

\begin{table}[htbp]
\centering
\footnotesize
\setlength{\tabcolsep}{6pt}
\begin{tabular}{@{}lcc@{}}
\toprule
Method & \shortstack{Activation\\coverage (\%)} & \shortstack{Boundary\\rejection (\%)} \\
\midrule
SAGE & 84.6 & 72.8 \\
Token Change & \textemdash & \textemdash \\
Claude Code, restricted$^{*}$ & 84.0 & 84.0 \\
Claude Code, unrestricted$^{*}$ & 97.6 & 90.7 \\
\midrule
DAFI & 91.6 & 91.2 \\
\bottomrule
\end{tabular}
\caption{Input-score component breakdown for the main baseline comparison. $^{*}$ denotes the 25-feature Claude Code evaluation; all other rows use 100 features. Dashes indicate inapplicable metrics.}
\label{tab:baseline-input-components}
\end{table}

Under unrestricted execution, Claude Code passes 19 features and consumes 332.39M tokens in total, or 13.296M per feature. Its median and maximum per-feature costs are 8.67M and 52.87M tokens, respectively, and it averages 194 assistant turns per feature. Table~\ref{tab:open-agent-comparison} summarizes both configurations.

\begin{table}[htbp]
\centering
\scriptsize
\setlength{\tabcolsep}{1.5pt}
\begin{tabular}{@{}lcccccc@{}}
\toprule
\multirow{2}{*}{Method} & \multirow{2}{*}{Joint passes} & \multicolumn{2}{c}{Input score (\%)} & \multirow{2}{*}{\shortstack{Output\\score (\%)}} & \multirow{2}{*}{\shortstack{Functional\\score}} & \multirow{2}{*}{\shortstack{Tokens / feature\\(M) $\downarrow$}} \\
\cmidrule(lr){3-4}
& & \shortstack{Activation\\coverage} & \shortstack{Boundary\\rejection} & & & \\
\midrule
Claude Code, restricted & 2/25 & 84.0 & 84.0 & 35.0 & 2.8 & 3.082 \\
Claude Code, unrestricted & 19/25 & 97.6 & 90.7 & 54.8 & 4.0 & 13.296 \\
\bottomrule
\end{tabular}
\caption{Results for the restricted and unrestricted Claude Code configurations on the same 25 features. Activation coverage, boundary rejection, and Output score are percentage-scaled; Output score is not a probability. Functional score uses a 1--5 scale, and costs are millions of tokens per feature. The restricted setting uses a round-start budget, not a hard cap.}
\label{tab:open-agent-comparison}
\end{table}

\section{Output Score Evolution and Output-Side Validation}
\label{sec:output-centric-validation}
Prior work on output-side feature descriptions argues that a faithful feature description should be evaluated through both the inputs that activate the feature and the outputs that change when the feature is stimulated. That framing directly motivates our Output score. Our difference is that we do not stop at an output-side description. We require the output evidence to support a functional interpretation from input context $X$ to output pattern $Y$, and we use the failure pattern across the three scores to decide what the agent should repair.

\subsection{Output Score Definition}
\label{sec:m2-definition}

For feature $i$, let $\Delta_i(v)$ denote the intervention-induced change in logit for token $v$ at the measured output position. The evaluator keeps the top positive-delta tokens
\begin{equation}
T_i^+ = \{v: \Delta_i(v) > 0\}
\end{equation}
and the top negative-delta tokens
\begin{equation}
T_i^- = \{v: \Delta_i(v) < 0\},
\end{equation}
after truncating each set to the configured top-$k$ evidence budget. An LLM judge is then given the output hypothesis $h_t^{\mathrm{out}}$ and the candidate tokens. It returns binary consistency indicators $a_v^+ \in \{0,1\}$ for each $v \in T_i^+$ and $a_v^- \in \{0,1\}$ for each $v \in T_i^-$. A positive token is marked consistent when it belongs to the output pattern described by $h_t^{\mathrm{out}}$. A negative token is marked consistent when the same hypothesis would also predict or cover that suppressed \mbox{token, which indicates that the hypothesis is too broad.}

When the total positive intervention mass exceeds $10^{-12}$, both coverage and penalty are normalized by this positive mass. The positive coverage term is the fraction of positive intervention mass explained by the hypothesis:
\begin{equation}
\mathrm{Coverage}(h_t^{\mathrm{out}},T_i^+)
=
\frac{\sum_{v\in T_i^+} a_v^+ \Delta_i(v)}
{\sum_{v\in T_i^+} \Delta_i(v)}.
\end{equation}
The negative penalty measures the suppressed-token mass covered by the hypothesis, normalized by the same total positive intervention mass:
\begin{equation}
\mathrm{Penalty}(h_t^{\mathrm{out}},T_i^-)
=
\frac{\sum_{v\in T_i^-} a_v^- |\Delta_i(v)|}
{\sum_{v\in T_i^+} \Delta_i(v)}.
\end{equation}
The output-effect metric is then
\begin{equation}
S_{\mathrm{out}}(i)
=
\mathrm{Coverage}(h_t^{\mathrm{out}},T_i^+)
-
\mathrm{Penalty}(h_t^{\mathrm{out}},T_i^-).
\end{equation}
This shared normalization expresses positive coverage and negative penalty on the same scale, making scores comparable across features whose raw intervention deltas have different scales. It also separates two failure modes: a low coverage score means that $h_t^{\mathrm{out}}$ misses the promoted output evidence, while a high penalty means that the hypothesis \mbox{covers tokens that the intervention actually suppresses.}

\paragraph{Relation to output-side baselines.}
We distinguish three output-side signals. VocabProj projects the SAE decoder direction through the unembedding matrix and is cheap, but it is a geometric proxy. Token Change-style evidence measures the tokens whose probabilities change after feature amplification, and is therefore closer to a causal output description. Output score follows the latter idea but ties the intervention to contexts where the feature already fires, because the dual-end target asks what the active feature does in its triggering context.

\begin{table}[htbp]
\centering
\footnotesize
\begin{tabular}{@{}lll@{}}
\toprule
Signal & Evidence & Role \\
\midrule
MaxAct & Activating texts & Input initialization \\
VocabProj & Top unembedded tokens & Output prior only \\
Token Change & Intervention deltas & Output evidence \\
Functional interpretation & $h_t^{\mathrm{in}}$, $h_t^{\mathrm{out}}$, $h_t^c$ & Final judgment \\
\bottomrule
\end{tabular}
\caption{Output-side signals used by related work and by our dual-end evaluator.}
\label{tab:output-signals}
\end{table}

\paragraph{Score evolution.}
The Output score went through three designs. The first design used the top 10 positive-delta tokens and rewarded the fraction covered by the proposed output hypothesis. This measured recall but was easy to game: a broad hypothesis could cover every candidate token and receive a high score even when the token set was semantically diffuse. The second design widened the positive evidence to the top 30 tokens and added negative-delta tokens. A selected negative token contributes its negative change, so over-broad hypotheses penalize themselves. The current design retains this precision pressure and calibrates \mbox{steering per feature to account for variation in logit changes.}

\begin{table}[htbp]
\centering
\footnotesize
\begin{tabular}{@{}lll@{}}
\toprule
Version & Scoring idea & Failure addressed \\
\midrule
V1 & Top positives & Broad-label recall \\
V2 & Positives and negatives & Vague output patterns \\
V3 & Per-feature calibration & Layer and scale drift \\
\bottomrule
\end{tabular}
\caption{Output-score revisions and the failures that motivated them.}
\label{tab:output-metric-evolution}
\end{table}

\paragraph{Layer bias of VocabProj.}
VocabProj is useful but not stable enough to serve as the output ground truth. In shallow layers, the projected tokens often resemble the input-side activating tokens, so projection can restate what made the feature fire. In deeper layers, projection becomes closer to the actual steering effect because the feature direction lies nearer to the logit-relevant subspace. The middle layers are the most informative for our method, because neither input evidence nor projection alone reliably describes the causal output effect. The Output score therefore uses intervention deltas.

\paragraph{Layer-wise consistency.}
Table~\ref{tab:m2-layerwise-token-change} reports Output scores on the same 100 matched features, with 20 features from each layer. DAFI achieves a higher score at every layer. We quantify consistency using the standard deviation across the five layer means. DAFI has an across-layer standard deviation of 0.042, compared with 0.123 for Token Change, indicating greater consistency across layers.

\begin{table}[htbp]
\centering
\footnotesize
\begin{tabular}{@{}lcc@{}}
\toprule
Layer & DAFI & Token Change \\
\midrule
L0 & 0.637 & 0.108 \\
L6 & 0.651 & 0.270 \\
L12 & 0.617 & 0.194 \\
L18 & 0.696 & 0.355 \\
L24 & 0.733 & 0.461 \\
\midrule
\textbf{Overall} & \textbf{0.667} & \textbf{0.278} \\
Across-layer SD $\downarrow$ & \textbf{0.042} & 0.123 \\
\bottomrule
\end{tabular}
\caption{Layer-wise Output score comparison between DAFI and Token Change on 100 matched features, with 20 features per layer. SD is the population standard deviation across the five layer means; lower values indicate more consistent performance across layers.}
\label{tab:m2-layerwise-token-change}
\end{table}

\section{Additional Case Studies}
\label{sec:additional-case-studies}
Section~\ref{sec:representative-cases} presents representative Shift, Equivalent, and Break cases. This appendix reports an additional case that illustrates refinement behavior.

\paragraph{L18 F6750, adding semantics.}
This feature initially passed none of the metrics. The base pipeline mixed narrow ``added'' semantics with broader supplement related tokens. After five agent rounds, the hypothesis narrowed to cross-lingual tokens expressing adding or having been added, including added, addition, append, and bolt on, while excluding adjacent concepts such as extra or supplementary. The final scores were Input score 1.000, Output score 0.934, and Functional score 5 out of 5.

\section{Threshold Robustness}
\label{sec:threshold-robustness}
Initial and adaptive explanations use the same scoring protocol. At the default thresholds, 20 of 250 initial explanations and 200 of 250 adaptive explanations satisfy the joint pass criterion.

Table~\ref{tab:threshold-sensitivity} varies one pass threshold at a time while holding the other two at their default values. The input-side pass criterion applies the same threshold to activation coverage and boundary rejection rather than thresholding their harmonic mean. The complete $3\times3\times3$ grid contains 27 operating points. Adaptive refinement raises the joint pass rate at all 27 points, by 6.4 to 72.0 percentage points.

\begin{table}[htbp]
\centering
\footnotesize
\setlength{\tabcolsep}{6pt}
\begin{tabular}{@{}lrrrr@{}}
\toprule
Varied metric & Threshold & Initial & Adaptive & Gain \\
\midrule
Input components & 0.7 & 8.0 & 80.0 & +72.0 \\
Input components & 0.8 & 8.0 & 80.0 & +72.0 \\
Input components & 0.9 & 6.4 & 49.6 & +43.2 \\
\midrule
Output score & 0.4 & 11.2 & 80.8 & +69.6 \\
Output score & 0.5 & 8.0 & 80.0 & +72.0 \\
Output score & 0.6 & 4.4 & 62.8 & +58.4 \\
\midrule
Functional score & 3 & 10.4 & 80.0 & +69.6 \\
Functional score & 4 & 8.0 & 80.0 & +72.0 \\
Functional score & 5 & 3.2 & 16.4 & +13.2 \\
\bottomrule
\end{tabular}
\caption{One-at-a-time threshold sensitivity over 250 features. The default input-component threshold is 0.8 for both activation coverage and boundary rejection; the default Output and Functional score thresholds are 0.5 and 4. Rates are percentages.}
\label{tab:threshold-sensitivity}
\end{table}

\section{Output Score Filter Analysis}
\label{sec:output-score-filter-analysis}

We evaluate the output-score filter of Arad et al. (2025) on Gemma-2-2B under the AxBench protocol and do not observe the gains they report on Gemma-2-9B under their own protocol. Model scale, judge model, instruction sample, and factor-selection rule all differ between the two setups, so the cross-study contrast cannot isolate a single cause; we list these differences to scope our conclusion.

\paragraph{Factor selection and decoding.}
Arad et al. choose the steering factor per feature to maximize a normalized ratio of generation success, how often the feature's top-20 logit-lens tokens appear in generated continuations, to perplexity measured with Gemma-2-9B, applying the rule to the five selection instructions at temperature 0.7 with factors up to 100. AxBench instead selects, on the same five instructions, the factor with the highest harmonic-mean LM-judge score over concept presence, instruction-following, and fluency; we follow this protocol with factors up to 10 and temperature-1.0 sampling without repetition penalties. In our setting the ratio rule selects factors yielding a mean holdout score of 0.105, against 0.140 for the AxBench rule on the same generations. Part of this gap is expected by construction, since the AxBench rule selects directly on the reported judge metric; we use it for the main comparison because it is the benchmark's official protocol and, applied identically to both filters, keeps the filter comparison controlled, with the ratio-rule run serving only as a transfer check on 2B. The remainder is plausibly a scale effect: the ratio rule retains factors up to 100, while in our 2B runs the mean holdout score declines beyond a factor of 2.0 and generations at factors of 3 and above are dominated by token repetition, consistent with the factor-versus-instruct tradeoff of Wu et al. (2025, Figure 4); we hypothesize that this \mbox{accounts for part of the gap between the two rules on 2B.}

\paragraph{Instructions.}
AxBench samples ten steering instructions per concept from AlpacaEval and does not release them, and Arad et al. draw their own genre-aligned samples. Their unfiltered 9B replication scores above the published AxBench SAE numbers, a gap they attribute partly to this sampling and partly to judge instability. We run the official AxBench codebase, so both filters are evaluated on the same seeded AlpacaEval sample. The sampling therefore confounds cross-study comparisons of absolute scores but not our matched-coverage comparison.

\paragraph{Judge model.}
Arad et al. score generations with Claude 3.7 Sonnet, whereas we follow AxBench and use gpt-4o-mini. Judges differ in their treatment of repetitive and partially steered text and can in principle reorder methods across studies, so the judge model is a further confound for cross-study comparison. Within our study, the same judge scores both filters \mbox{on identical generations, controlling for this difference.}

\paragraph{Model scale.}
The Output score is computed via a causal intervention that measures rank-weighted probability shifts for top tokens, and the gains of Arad et al. are reported on the 9B model. On layer 20 of Gemma-2-2B we find that high Output scores co-occur less reliably with generations that remain fluent and instruction-following, so the filter retains pairs that the AxBench judge scores low. We do not re-run the ratio rule on 9B, so our results do not speak to the effectiveness of the filter under the original 9B protocol; they show that the \mbox{filter does not transfer to 2B under the AxBench protocol.}

Taken together, these differences indicate that absolute steering scores on Concept500 are sensitive to protocol details, and that our Output score null result on 2B is consistent with, rather than a disproof of, the positive 9B result of Arad et al. Our conclusions rely on comparing filters on identical generations under one protocol, controlling for these differences.

\section{Reproducibility Details}
Strict success uses fixed thresholds throughout the reported comparisons. The reported Input score is the harmonic mean of activation coverage and boundary rejection, but the input-side pass criterion remains stricter: both components must be at least 0.8. Output score requires a score of at least 0.5, and Functional score requires a score of at least 4. The candidate pool and round budget are fixed before evaluation. Candidate selection retains one complete, internally consistent interpretation triplet with its associated scores, with ties resolved by the earlier round. All three reported scores refer to this same triplet, and joint success requires it to pass all three criteria. Intervention keeps the top 30 positive and negative token changes, uses at most five steering prompts, and targets the maximally activating token by default. The maximum activation scale is 2.0, with a last-token scale of 1.0 when that scope is selected. The GemmaScope study comprises 200 features interpreted during skill accumulation and 50 held-out validation features, with 40 and 10 features, respectively, from each of layers 0, 6, 12, 18, and 24. We reuse the checkpoint-120 validation results in the final 250-feature analysis to avoid an additional evaluation run. The 100-feature subsets used for the component-matched baseline comparison and the short-context-probing comparison were sampled from this 250-feature pool. The Qwen-Scope transfer study fixes 20 features in each of layers 0, 7, 14, 21, and 27. DeepSeek-V4-Pro is the language-model backbone for explanation generation and judging, and tool calls use temperature 0. Within each comparison, the evaluated methods use the same feature list and the applicable shared evaluation settings; comparison-specific differences are described in their corresponding experimental protocols. Token counts include the \mbox{complete agent context under each evaluated interface.}

\section{Ethics Considerations}
The human evaluation was conducted by two authors with experience in SAE-based interpretability. They independently rated the same model-generated feature interpretations while blinded to the system scores and to each other’s labels. The evaluation involved no recruitment of external participants, no additional compensation, and no collection of personal, sensitive, or behavioral data. Language models are used to generate and judge feature explanations; their outputs may inherit model biases and should therefore be interpreted in conjunction with empirical evidence and human review.

\end{CJK*}
\end{document}